\pdfoutput=1
\documentclass[10pt]{article}
\usepackage{fullpage}
\usepackage[margin=1in]{geometry}
\usepackage{graphicx}
\usepackage{booktabs}

\usepackage[utf8]{inputenc} 

\usepackage{natbib}

\usepackage[colorlinks,linkcolor={blue}, citecolor={blue}, urlcolor={blue}, linktoc=all]{hyperref}

\usepackage{xcolor}

\usepackage{multirow}
\usepackage{array}
\usepackage{setspace}
\usepackage{float}
\usepackage[font=small]{caption}
\usepackage{hhline}

\usepackage{adjustbox}

\usepackage[format=hang]{subcaption}

\usepackage{amsthm}
\usepackage{thmtools}
\usepackage{mathtools}
\usepackage{amsmath}
\usepackage{enumitem}
\usepackage{amsfonts}
\usepackage{graphicx}
\usepackage{comment}
\usepackage{algorithmic}
\usepackage[noline,ruled, noend]{algorithm2e}
\usepackage{setspace}
\usepackage{amssymb}

\newtheorem{theorem}{Theorem}
\newtheorem{lemma}{Lemma}

\newtheorem{corollary}[lemma]{Corollary}
\newtheorem{definition}[lemma]{Definition}
\renewenvironment{proof}[1][Proof]{\begin{trivlist}
\item[\hskip \labelsep {\bfseries #1}]}{\qed\end{trivlist}}

\newcommand*\R[0]{\mathbb{R}}

\newcommand*\E[1]{\mathbb{E}\left[#1\right]}

\newcommand*\F[0]{\mathcal{F}}

\newcommand\numberthis{\addtocounter{equation}{1}\tag{\theequation}}
\newcommand*\lrb[1]{\left[#1\right]}

\newcommand*\lrbb[1]{\left\{#1\right\}}
\newcommand*\lrp[1]{\left(#1\right)}
\newcommand*\lrn[1]{\left\|#1\right\|}

\renewcommand*{\qed}{\hfill\ensuremath{\blacksquare}}

\newcommand*\expectation[0]{\mathbb{E}}

\newcommand*\reals{\mathbb{R}}
\newcommand*\states{\mathcal{S}}
\newcommand*\actions{\mathcal{A}}

\def\rT{\mathrm{T}}

\newcounter{assumption}%
\renewcommand{\theassumption}{\arabic{assumption}}

\newenvironment{assumption}[1][]{\begin{trivlist}\item[] \refstepcounter{assumption}%
 {\bf{Assumption\ \theassumption\ }}{(#1)}.  }{%
 \ifvmode\smallskip\fi\end{trivlist}}
 
 \newcommand{\Ind}{\ensuremath{\mathbb{I}}}

\newcommand{\Prob}{\ensuremath{{\mathbb{P}}}}

\DeclareMathOperator*{\argmax}{argmax} 

\newcommand{\Nikki}[1]{\textcolor{red}{[Nikki: #1]}}

\title{Langevin Thompson Sampling with Logarithmic Communication: Bandits and Reinforcement Learning}

\author{
    \begin{tabular}{cc}
    \shortstack{Amin Karbasi\\ \\
      Yale University\\ 
      \texttt{amin.karbasi@yale.edu}}
      &
      \shortstack{Nikki Lijing Kuang\footnotemark[1]\thanks{Major contribution credited equally to Kuang and Mitra.}\\ \\
      University of California, San Diego\\ 
      \texttt{l1kuang@ucsd.edu}} 
      \\ \\
      \shortstack{Yi-An Ma\\ \\
      University of California, San Diego\\ 
      \texttt{yianma@ucsd.edu}}
      &
      \shortstack{Siddharth Mitra\footnotemark[1]\\ \\
      Yale University\\ 
      \texttt{siddharth.mitra@yale.edu}}
    \end{tabular}
  }

\date{}

\begin{document}










\maketitle

\begin{abstract}
Thompson sampling (TS) is widely used in sequential decision making due to its ease of use and appealing empirical performance. However, many existing analytical and empirical results for TS rely on restrictive assumptions on reward distributions, such as belonging to conjugate families, which limits their applicability in realistic scenarios. Moreover, sequential decision making problems are often carried out in a batched manner, either due to the inherent nature of the problem or to serve the purpose of reducing communication and computation costs. In this work, we jointly study these problems in two popular settings, namely, stochastic multi-armed bandits (MABs) and infinite-horizon reinforcement learning (RL), where TS is used to learn the unknown reward distributions and transition dynamics, respectively. We propose batched \textit{Langevin Thompson Sampling} algorithms that leverage MCMC methods to sample from approximate posteriors with only logarithmic communication costs in terms of batches.
Our algorithms are computationally efficient and maintain the same order-optimal regret guarantees of $\mathcal{O}(\log T)$ for stochastic MABs, and $\mathcal{O}(\sqrt{T})$ for RL. We complement our theoretical findings with experimental results.
\end{abstract}

\section{Introduction}

\begin{table*}[t]
    \def\arraystretch{1.5}
    \centering
    \small
    \begin{tabular}{|>{\centering\arraybackslash}m{1.7cm}
               |>{\centering\arraybackslash}m{3.6cm}
               |>{\centering\arraybackslash}m{2.2cm}
               |>{\centering\arraybackslash}m{2cm}
                |>{\centering\arraybackslash}m{1.5cm}
                |>{\centering\arraybackslash}m{2.2 cm}|}
\hline 
  & \textbf{TS-based Algorithms} & \textbf{Batching Scheme} & \textbf{MCMC Method} & \textbf{Regret} & \textbf{\# of Batches} \\
 \hline
 \multirow{3}{1.7cm}{\centering Stochastic MAB} & \cite{karbasi2021parallelizing} & Dynamic & - & $O(\log T)$ & $O(\log T)$ \\
 \cline{2-6}
 & \cite{mazumdar2020thompson} & - & SGLD, ULA & $O(\log T)$ & ${O}(T)$ \\
 \cline{2-6}
 & This paper (Algorithm \ref{alg:TS_Bandit}) & Dynamic & SGLD & ${O}(\log T)$ & ${O}(\log T)$\\
 \hhline{|=|=|=|=|=|=|}
 \multirow{4}{1.7cm}{\centering TS for RL (PSRL)} & \cite{osband2013more} & - & - & ${O}(\sqrt {T})$ & ${O}(T/H)$ \\
 \cline{2-6}
 & \cite{ouyang2017learning} & Dynamic & - & ${O}(\sqrt {T})$ & ${O}(\sqrt{T})$ \\
 \cline{2-6}
 & \cite{theocharous2017posterior} & Static & - & ${O}(\sqrt{T})$ & ${O}(\log T)$ \\
 \cline{2-6}
 & This paper (Algorithm \ref{alg:TS_alg_MDP_double}) & Static & SGLD, MLD  & ${O}(\sqrt {T})$ & ${O}(\log T)$\\
 \hline
    \end{tabular}
    \vspace{-0.2cm}
    \caption{We compare our methods with existing TS-based algorithms in terms of batching schemes and MCMC methods adopted for approximation. Performance is measured by regret, while communication cost is quantified with the number of batches. Here, $T$ is the time horizon, $H$ is the fixed episode length in episodic MDP settings. Our methods achieve optimal performance, while reducing computation and communication costs due to batching, and are applicable in broader regimes.} 
    \label{tab:result_comparison}
\end{table*}

Modern machine learning often needs to balance computation and communication budgets with statistical guarantees.
Existing analyses of sequential decision making have been primarily focused on the statistical aspects of the problems \citep{tian2020towards}, less is known about their computation and communication aspects. In particular, regret minimization in multi-armed bandits (MABs) and reinforcement learning (RL) \citep{jaksch2010near, wu2022nearly,  jung2019thompson} is often studied under the common assumptions that computation can be performed perfectly in time, and that communication always happens in real-time \citep{li2022stochastic, jin2018q, haarnoja2018soft}.

A question of particular importance is whether optimal decisions can still be made under reasonable computation and communication budgets. In this work, we study the exploration-exploitation problem with low computation and communication costs using Thompson Sampling (TS) \citep{thompson1933likelihood} (a.k.a. posterior sampling). To allow sampling from distributions that deviate from the standard restrictive assumptions, and to enable its deployment in settings where computing the exact posterior is challenging, we employ Markov Chain Monte Carlo (MCMC) methods. 

TS operates by maintaining a posterior distribution over the unknown and is widely used owing to its strong empirical performance \cite{chapelle2011empirical}. However, the theoretical understanding of TS in bandits typically relied on the restrictive conjugacy assumptions between priors and reward distributions. Recently, approximate TS methods for general posterior distributions start to be studied \cite{mazumdar2020thompson, xu2022langevin}, where MCMC algorithms are used in conjunction with TS to expand its applicability in fully-sequential settings. On the other hand, to the best of our knowledge, how to provably incorporate MCMC with posterior sampling in RL domains remains to be untackled.

Moreover, previous analyses of TS have been restricted to the fully-sequential settings, where feedback or reward is assumed to be immediately observable upon taking actions (i.e. before making the next decision). Nevertheless, it fails to account for the practical settings when delays take place in communication, or when feedback is only available in an aggregate or batched manner. Examples include clinical trials where feedback about the efficacy of medication is only available after a nontrivial amount of time,  recommender systems where feedback from multiple users comes all at once and marketing campaigns \cite{schwartz2017customer}. This issue has been studied in literature by considering static or dynamic batching schemes and by designing algorithms that acquire feedback in batches, where the learner typically receives reward information only at the end of the batch \citep{karbasi2021parallelizing,kalkanli2021batched, vernade2020linear, zhang20neural}. Nonetheless, the analysis of approximate TS in batched settings is unavailable for both bandits and RL.

In this paper, we tackle these challenges by incorporating TS with Langevin Monte Carlo (LMC) and batching schemes in stochastic MABs and infinite-horizon RL. Our algorithms are applicable to a broad class of distributions with only logarithmic rounds of communication between the learner and the environment, thus being robust to constraints on communication. We compare our results with other works in Table \ref{tab:result_comparison}, and summarize our main contributions as follows:

\vspace{-0.2cm}
\begin{itemize}
\setlength\itemsep{0.2em}

\item For stochastic MABs with time horizon $T$, we present \textit{Langevin Thompson Sampling} (BLTS, Algorithm \ref{alg:TS_Bandit}) along with Theorem \ref{thm:reg_batch_SGLD}, which achieves the optimal $ O(\log T)$ regret with $ O(\log T)$ batches\footnote{A $T$ round game can be thought of as $T$ many batches each of size 1.}. The main technical contribution here is to show that when feedback is obtained in a batched manner where the posterior concentration is weaker (Theorem \ref{thm:sgld_convergence}), the convergence guarantee of SGLD continues to hold.

\item For large-scale infinite-horizon MDPs, we present \textit{Langevin Posterior Sampling for RL} (LPSRL, Algorithm \ref{alg:TS_alg_MDP_double}) along with Theorem \ref{thm:mdp_general} to show that SGLD with a static policy-switching\footnote{In MDP settings, the notion of a batch is more appropriately thought of as a policy-switch.} scheme achieves the optimal $ O(\sqrt{T} )$ Bayesian regret with $ O(\log T)$ policy switches. 
For tabular MDPs, we show that LPSRL incorporated with the Mirrored Langevin Dynamics (MLD) achieves the optimal $ O(\sqrt{T} )$ Bayesian regret with $ O(\log T)$ policy switches.
The use of approximate sampling leads to an additive error where the true model and the sampled model are no longer identically distributed. This error can be properly handled with the convergence guarantees of LMC methods.

\item Experiments are performed to demonstrate the effectiveness of our algorithms, which maintain the order-optimal regret with significantly lower communication costs compared to existing exact TS methods.



\end{itemize}

\section{Problem Setting} \label{sec:problem_setting}

In this section, we introduce the problem setting with relevant background information.

\subsection{Stochastic Multi-armed Bandits}
\label{sec:pre_bandit}

We consider the $N$-armed stochastic multi-armed bandit problem, where the set of arms is denoted by $\mathcal{A} = [N] = \{1,2,\dots, N\}$. Let $T$ be the time horizon of the game. At $t = 1,2,\dots, T$, the learner chooses an arm $a_t\in \mathcal{A}$ and receives a real-valued reward $r_{a_t}$ drawn from a fixed, unknown, parametric distribution corresponding to arm $a_t$. In the standard fully-sequential setup, the learner observes rewards immediately. Here, we consider the more general batched setting, where the learner observes the rewards for all timesteps within a batch at the end of it. We use $B_k$ to denote the starting time of the $k$-th batch, $B(t)$ to represent the starting time of the batch that contains time $t$, and $K$ as the total number of batches. The learner observes the set of rewards $\{ r_{a_t}\}_{t = B_k}^{B_{k+1}-1}$ at the end of the $k$-th batch. Note that the batched setting reduces to the fully-sequential setting when the number of batches is $T$, each with size $1$.  

Suppose for each arm $a$, there exists a parametric reward distribution parameterized by $\theta_a \in \R^d$ such that the true reward distribution is given by $p_a(r) = p_a(r | \theta^*_a)$, where $\theta^*_a$ is an unknown parameter\footnote{Our results hold for the more general case of $\theta_a \in \R^{d_a}$, but for simplicity of exposition, we consider the ambient dimension for the parameters of each arm to be the same.}. To ensure meaningful results, we impose the following assumptions on the reward distributions for all $a \in \mathcal{A}$:
\vspace{-0.2cm}
\begin{itemize}
\setlength\itemsep{0.2em}
    \item \textbf{Assumption 1:} $\log p_a(r|\theta_a)$ is $L$-smooth and $m$-strongly concave in $\theta_a$.
    \item \textbf{Assumption 2:} $ p_a(r|\theta_a^*)$ is $\nu$ strongly log-concave in $r$ and $\nabla_\theta \log p_a (r|\theta^*_a)$ is $L$-Lipschitz in $r$.
    \item \textbf{Assumption 3:} The prior $\lambda_a (\theta_a)$ is concave with $L$-Lipschitz gradients for all $\theta_a$.
    \item \textbf{Assumption 4:} Joint Lipschitz smoothness of (the bivariate) $\log p_a(r | \theta_a)$ in $r$ and $\theta_a$.
\end{itemize}
These properties include log-concavity and Lipschitz smoothness of the parametric families and prior distributions, which are standard assumptions in existing literature \citet{mazumdar2020thompson} and are satisfied by models like Gaussian bandits \citep{honda13gaussian}. For the sake of brevity, We provide the mathematical statements of these assumptions in Appendix \ref{app:assumptions}.

Let $\mu_a$ denote the expected value of the true reward distribution for arm $a$. The goal of the learner is to minimize the expected regret, which is defined as follows: 
\begin{equation}
    R (T) := \E{\sum_{t=1}^T \mu^* - \mu_{a_t}} = \sum_{a \in \mathcal{A} } \Delta_a \E{k_a(T)}, 
\end{equation}

where $\mu^* = \max_{a\in \mathcal{A}} \mu_a$, $\Delta_a = \mu^* - \mu_a$, and $k_a(t)$ represents the number of times arm $a$ has been played up to time $t$. Without loss of generality, we will assume that arm $1$ is the best arm. We discuss the MAB setting in Section \ref{sec:bandits_main}.

\subsection{Infinite-horizon Markov Decision Processes}
\label{sec:pre_mdp}
We focus on average-reward MDPs with infinite horizon \citep{jaksch2010near, wei2021learning}, which is underexplored compared to the episodic setting. It is a more realistic model for real-world tasks, such as robotics and financial-market decision making, where state reset is not possible. Specifically, we consider an undiscounted \textit{weakly communicating} MDP $(\mathcal{S}, \mathcal{A}, p, \mathcal{R})$ with  infinite time horizon\footnote{It is known that weakly communicating MDPs satisfy the Bellman Optimality.} \citep{ouyang2017learning, theocharous2017posterior}, where $\mathcal{S}$ is the state space, $\mathcal{A}$ is the action space, $p$ represents the parameterized transition dynamics, and $\mathcal{R}: \mathcal{S}\times \mathcal{A}$ $ \to \R$ is the reward function. We assume that $\theta \in \R^d$ parameterizes the transition dynamics and there exists a true unknown $\theta^*$ governing the next state of the learner. At each time $t$, the learner is in state $s_t$, takes action $a_t$, and transits into the next state $s_{t+1}$, which is drawn from $p(\cdot | s_t, a_t, \theta^*)$. 

We consider two sub-settings based on the parameterization of the transition dynamics: the General Parameterization and the Simplex Parameterization. These sub-settings require different assumptions and setups, which we elaborate on in their respective sections (Section \ref{sec:general_mdp} and Section \ref{sec:tabular_mdp}). In the MDP context, the notion of batch is more appropriately thought of as a policy switch. Therefore, $B_k$ now represents the starting time of the $k$-th policy switch, and we additionally define $T_k$ as the number of time steps between policy switch $k$ and $k+1$. 
We consider stationary and deterministic policies, which are mappings from $\mathcal{S} \to \mathcal{A}$. Let $\pi_k$ be the policy followed by the learner after the $k$-th policy switch.
When the decision to update and obtain the $k$-th policy is made, the learner uses the observed data $\{s_t, a_t, \mathcal{R}(s_t, a_t), s_{t+1}\}_{t=B_k}^{B_{k+1} -1}$ collected after the $(k-1)$-th policy switch to sample from the updated posterior and compute $\pi_k$.
The goal of the learner is to maximize the long-term average reward: $$J^{\pi}(\theta) =  \E{\limsup_{T \rightarrow \infty} \frac{1}{T} \sum_{t=1}^T \mathcal{R}(s_t, a_t) }.$$ Similar to other works \cite{ouyang2017learning, osband2013more, theocharous2017posterior}, we measure the performance using Bayesian regret\footnote{In Bayesian regret, expectation is taken $w.r.t$ the prior distribution of the true parameter $\theta^*$,
the randomness of algorithm and transition dynamics.} defined by: 
\begin{equation}
    R_B (T) := \E{\sum_{t=1}^T (J^{\pi^*}(\theta^*) - \mathcal{R}(s_t, a_t))},
\end{equation} where $J^{\pi^*}(\theta^*)$ denotes the average long-term reward after running the optimal policy under the true model.

It is known that weakly communicating MDPs satisfy the following Bellman optimality \cite{bertsekas2012dynamic, ouyang2017learning, wei2021learning} in infinite-horizon setting, and there exists some positive number $H$ such that the span (Definition~\ref{def:span}) satisfies $sp(h(\theta)) \leq H$ for all $\theta \in \R^d$.

\begin{lemma} [Bellman Optimality]
\label{lem:bellman_mdp}
There exist optimal average reward $J \in \R$ and a bounded measurable function $h: \mathcal{S} \rightarrow \R$, such that for any $s, \in \mathcal{S}, \theta \in \R^d$, the Bellman optimality equation holds:
\begin{align}
\label{eqn:mdp_bellman}
    J(\theta) + h(s, \theta) = \max\limits_{a \in \mathcal{A}} \Big\{ \mathcal{R}(s, a) + \mathbb{E}_{s^\prime \sim p(\cdot | s, a; \theta)}[h(s^\prime, \theta)] \Big\}.
\end{align}
Here $J(\theta) = max_{\pi}J^{\pi}(\theta)$ under $\theta$ and is independent of initial state. Function $h^\pi(s, \theta) = \lim_{T \rightarrow \infty}\mathbb{E}[\sum_{t=1}^T$ $(\mathcal{R}(s_t, \pi(s_t)) - J^\pi(s_t)) | s_1 = s]$ quantifies the bias of policy $\pi$ $w.r.t$ the average-reward under $\theta$, and $h(s, \theta) = h^{\pi^*}(s, \theta)$, where ${\pi^*} = \argmax_{\pi} J^{\pi}(\theta)$.
\end{lemma}
\begin{definition}
\label{def:span}
    For any $\theta \in \R^d$, span of an MDP is defined as $sp(h(\theta)) := sup_{s, s^\prime \in \mathcal{S}} |h(s, \theta) - h(s^\prime, \theta) | = max_{s \in \mathcal{S}} h(s, \theta) - min_{s \in \mathcal{S}} h(s, \theta)$.
\end{definition} 

\section{Related Work} \label{sec:related_works}

Under the conjugacy assumptions on rewards, asymptotic convergence of TS was studied in stochastic MABs by \citet{granmo2010solving} and \citet{may2012optimistic}. Later, finite-time analyses with $O(\log T)$ problem-dependent regret bound were provided \cite{agrawal2012analysis, kaufmann2012thompson, agrawal2013further}. However, in practice, exact posteriors are intractable for all but the simplest models \citep{riquelme2018deep}, necessitating the use of approximate sampling methods with TS in complex problem domains. Recent progress has been made in understanding approximate TS in fully-sequential MABs \cite{lu2017ensemble, mazumdar2020thompson, zhang2022feel, xu2022langevin}. On the other hand, the question of learning with TS in the presence of batched data has evolved along a separate trajectory of works \citep{karbasi2021parallelizing,kalkanli2021batched, vernade2020linear, zhang20neural}. However, provably performing Langevin TS in batched settings remains unexplored, and in this paper, we aim at bridging these lines.

Moving to the more complex decision-making frameworks based on MDPs, TS is employed in model-based methods to learn transition models, which is known as \textit{Posterior Sampling for Reinforcement Learning} (PSRL) \citep{strens2000bayesian}. When exact posteriors are intractable, MCMC methods have been empirically studied for performing Bayesian inference in policy and reward spaces in RL \citep{brown2020safe, imani2018bayesian, bojun2020steady, guez2014better}. MCMC is a family of approximate posterior inference methods that enables sampling without exact knowledge of posteriors \cite{ma2015complete, welling2011bayesian}. However, it is unclear how to provably incorporate MCMC methods in learning transition models for RL.


Furthermore, the analysis of undiscounted infinite-horizon MDPs \citep{yadkori15bayesian, osband2016posterior, ouyang2017learning, wei2020model, wei2021learning} poses greater challenges compared to the well-studied episodic MDPs with finite horizon and fixed episode length \citep{osband2013more}. Previous works on infinite-horizon settings include model-based methods that estimate environment dynamics and switch policies when the number of visits to state-action pairs doubles \citep{jaksch2010near, tossou2019near, agrawal2017optimistic, bartlett2012regal}. Nevertheless, under such dynamic schemes, the number of policy switches can be as large as $O(\sqrt{T})$, making it computationally heavy and infeasible for continuous states and actions. To enable TS with logarithmic policy switches while maintaining optimal regret, we build upon an algorithmically-independent static scheme as in \citet{theocharous2017posterior}, and incorporate Langevin Monte Carlo (LMC) methods to sample from inexact posteriors.

\section{SGLD for Langevin Thompson sampling}
\label{sec:pre_TS_SGLD}

In the MAB and MDP settings, $\theta$ parameterizes the unknown reward or transition distributions respectively. TS maintains a distribution over the parameters and updates the distribution to (the new) posterior upon receiving new data. Given $p(X|\theta)$, prior $\lambda (\theta)$, and $n$ data samples $\{X_i\}_{i=1}^n$
\footnote{Here data $\{X_i\}_{i=1}^n$ can be rewards for some arms or actual transitions of state-action pairs depending on the setting.}, let $\rho_n$ be the posterior distribution after receiving $n$ data samples which satisfies: $\rho(\theta | \{X_i\}_{i=1}^n) \propto \exp (\sum_{i=1}^n \log p(X_i | \theta)  + \log \lambda (\theta))$. In addition, consider the scaled posterior $\rho_n [\gamma]$ for some scaling parameter $\gamma$, which represents the density proportional to $\exp (\gamma(\sum_{i=1}^n \log p(X_i | \theta) + \log \lambda (\theta)))$. 

The introduction of MCMC methods arises from the need for sampling from intractable posteriors in the absence of conjugacy assumptions. We resort to a gradient-based MCMC method that performs noisy updates based on Langevin dynamics: Stochastic Gradient Langevin Dynamics (SGLD). Algorithm \ref{alg:TS_SGLD} presents SGLD with bached data to generate samples from an approximation of the true posterior. For a detailed exposition, please refer to \cite{welling2011bayesian, ma2015complete} and Appendix \ref{app:MCMC}. Algorithm \ref{alg:TS_SGLD} takes all available data $\{X_s\}_{s=1}^n$ at the start of a batch $b$ as input, subsamples data, performs gradient updates by computing $\nabla \widehat{U}(\theta) = -\frac{n}{|D|}\sum_{X_s \in D} \nabla \log p(X_s | \theta) - \nabla \log \lambda (\theta)$, and outputs the posterior for batch $b$.

\IncMargin{1em}
\begin{algorithm}[ht]
	\setstretch{0.8}
	\SetAlgoLined
	\KwIn{prior $\lambda(\theta)$, data $\{X_s\}_{s=1}^n$, sample from last batch $\theta^{b-1}$, total iterations $N$, learning rate $\eta$, parameters $L$, scaling parameter $\gamma$.}

	\textbf{Initialization}: 
	$\theta_0 \leftarrow \theta^{b-1}$ \\
	        \For{$i = 1, \dots, N$}{
	            Subsample $D \subseteq \{X_s\}_{s=1}^n$ \\
                Compute $\nabla \widehat{U}(\theta_{i\eta})$ over $D$\\
	            Sample $\theta_{(i+1)\eta} \sim \mathcal{N}(\theta_{i\eta} - \eta \nabla \widehat{U}(\theta_{i\eta}),\ 2 \eta I )$
	        }
	       \KwOut{ $\theta^b \sim  \mathcal{N} \big(\theta_{N\eta},\ \frac{1}{n L\gamma } I \big)$}
	\caption{SGLD with Batched Data}
	\label{alg:TS_SGLD}
\end{algorithm}
\DecMargin{1em}

In the batched setting, new data is received at the end of a batch, or when making the decision to perform a new policy switch. Due to the way that the learner receives new data and the fact that the batch data size may increase exponentially\footnote{Data received in batch $k$ can be doubled compared to the previous batch.}, the posterior concentrates slower. This differs from the fully-sequential problem where the distribution shift of successive true posteriors is small owing to data being received in an iterative manner.  We show that in batched settings, with only constant computational complexity in terms of iterations, SGLD is able to provide strong convergence guarantee as in the fully-sequential setting \cite{mazumdar2020thompson}. Theorem \ref{thm:sgld_convergence} shows the convergence of SGLD in the Wasserstein-$p$ distance can be achieved with a constant number of iterations and data.

\begin{theorem} [SGLD convergence]
\label{thm:sgld_convergence}

Suppose that the parametric reward/transition families, priors, and true reward/transition distributions satisfy Assumptions \ref{ass:1}-\ref{ass:4}. Let $\kappa := \max \{L / m, L / \nu\}$, $|D| = O(\kappa^2)$, $\eta = O(1/n\kappa L)$, and $N = O(\kappa^2)$, then for any $\delta \in (0,1)$, the following holds with probability $\geq 1-\delta$:
\[
W_p \left(\tilde{\rho}_{n}, \rho_{n}\right) \leq \sqrt{\frac{12}{nm}}(d + \log Q + (32 + 8d\kappa^2)p)^{1/2}
\]
for all $p \geq 2$, and where $Q := \max_{\theta}\frac{ \lambda(\theta)}{\lambda(\theta^*)}$ measures the quality of prior distribution.

\end{theorem}

$\rho_n$ denotes the true posterior corresponding to $n$ data samples and $\Tilde{\rho}_n$ is the approximate posterior outputted by Algorithm \ref{alg:TS_SGLD}. 
We also note that similar concentration bounds can be achieved by using the \textit{Unadjusted Langevin Algorithm (ULA)} for batched data, which adopts full-batch gradient evaluations and therefore leads to a growing iteration complexity. The proofs of Theorem~\ref{thm:sgld_convergence} are adapted to the batched setting, which differs from \cite{mazumdar2020thompson}.

\vspace{-0.2cm}
\section{Batched Langevin Thompson Sampling for Bandits}
\label{sec:bandits_main}

In this section, we introduce \textit{Langevin Thompson Sampling} for batched stochastic MAB setting in Algorithm \ref{alg:TS_Bandit}, namely, BLTS. It leverages SGLD and batching schemes to learn a wide class of unknown reward distributions while reducing communication and computation costs. We have previously discussed the results of SGLD in Section \ref{sec:pre_TS_SGLD} for both MABs and MDPs.  Here, we focus on the batching strategy in Algorithm \ref{alg:TS_Bandit} for bandits, and discuss the resulting regret guarantee.

\subsection{Dynamic Doubling Batching Scheme}

BLTS keeps track of the number of times each arm $a$ has been played until time $t$ with $k_a (t)$. Initially, all $\{k_a\}_{a \in \mathcal{A}}$ are set to 0. The size of each batch is determined by $\{k_a\}_{a \in \mathcal{A}}$ and the corresponding integers $\{l_a\}_{a \in \mathcal{A}}$. Once $k_a$ reaches $2^{l_a}$for some arm $a$, BLTS makes the decision to terminate the current batch, collects all rewards from the batch in a single request, and increases $l_a$ by $1$. BLTS thus starts a new batch whenever an arm is played twice as many times as in the previous batch, which results in growing batch sizes. As the decision to move onto the next batch depends on the sequence of arms that is played, it is considered as ``dynamic". This batching scheme is similar to the one used in \citet{karbasi2021parallelizing}. The total number of batches that BLTS carries out satisfies the following theorem, and its proof can be found in Appendix \ref{app:bandit}.

\begin{restatable}{theorem}{banditBatches}
\label{thm:bandit_log_batch}
BLTS ensures that the total number of batches is at most $O(N \log T)$ where $N = |\mathcal{A}|$.
\end{restatable}

\citet{gao19batched} showed that $\Omega (\log T/ \log \log T)$ batches are required to achieve the optimal logarithmic dependence in time horizon $T$ for a batched MAB problem. This shows that the dependence on $T$ in the number of batches BLTS requires is at most a factor of $\log \log T $ off the optimal.
We now state and discuss the BLTS algorithm.

\subsection{Regret of BLTS Algorithm}

In Algorithm \ref{alg:TS_Bandit}, denote by $\theta_a^k$ the output of Algorithm \ref{alg:TS_SGLD} for arm $a$ at batch $k$. At the end of each batch, new data is acquired all at once and the posterior is being updated. It is important to note that upon receiving new data when we run Algorithm \ref{alg:TS_SGLD} for each arm, only that arm's data is fed into Algorithm \ref{alg:TS_SGLD}. For each $a \in \mathcal{A}$, assume the existence of linear map $\alpha_a$ such that $\mathbb{E}_{X \sim p_a(X|\theta_a)} [X] = \alpha_a^\intercal ~ \theta_a~ \forall \theta_a \in \R^d$, where $\lrn{\alpha_a}$ is bounded. Theorem \ref{thm:reg_batch_SGLD} shows the regret guarantee of BLTS.

\begin{restatable}{theorem}{banditRegret}
\label{thm:reg_batch_SGLD}

Assume that the parametric reward families, priors, and true reward distributions satisfy Assumptions 1 through 4 for each arm $a \in \mathcal{A}$. Then with the SGLD parameters specified as per Algorithm \ref{alg:TS_SGLD} and with $\gamma = O(1/d\kappa^3)$ (for $\kappa := \max \{L / m, L / \nu\}$), BLTS satisfies:
\begin{align*}
     R(T) &\leq \sum_{a > 1} \frac{C \sqrt{Q_1} }{m\Delta_a} \lrp{d + \log Q_1 + d\kappa^2 \log T + d^2 \kappa^2 } + \frac{C}{m \Delta_a} \lrp{d + \log Q_a +  d^2 \kappa^2 \log T } + 4\Delta_a,
 \end{align*}
where $C$ is a constant and $Q_a := \max_{\theta}\frac{ \lambda_a(\theta)}{\lambda_a(\theta^*)}$. The total number of SGLD iterations used by BLTS is $O(\kappa^2NlogT)$.
\end{restatable}

\paragraph{Discussion} We show that BLTS achieves the optimal $O\Big(\frac{\log T}{\triangle}\Big)$ regret bound with exponentially fewer rounds of communication between the learner and the environment. Result of Theorem~\ref{thm:reg_batch_SGLD} relies on both the statistical guarantee provided by SGLD and the design of our batching scheme. In bached setting, one must carefully consider the trade-off between batch size and the number of batches. While it is desirable to reuse the existing posterior for sampling within a batch, the batching scheme must also ensure
new data is collected in time to
avoid significant distribution shifts. In addition, the use of SGLD allows BLTS to be applicable in a wide range of general settings with a low computation cost of $O(\kappa^2N\log T)$.

In the regret bound of Theorem~\ref{thm:reg_batch_SGLD}, $Q_a$  measures the quality of prior for arm $a$. Specifically, if the prior is properly centered such that its mode is at $\theta_a^*$, or if the prior is uninformative or flat everywhere, then $\log Q_a = 0$.  In Section~\ref{sec:empirical}, we show that using either favorable priors or uninformative priors provides similar empirical performance as existing methods.  

\vspace{-0.1cm}
\IncMargin{1em}
\begin{algorithm}[!ht]
	\setstretch{0.8}
	\SetAlgoLined
	\KwIn{priors $\lambda_a(\theta)~ \forall a \in \mathcal{A}$, scaling parameter $\gamma$, inputs for SGLD subroutine $N, \eta, L$. }

	\textbf{Initialization}: $k_a \leftarrow 0, l_a \leftarrow 0, n_a \leftarrow 0, \Tilde{\rho}_{a,k} = \Tilde{\rho}_{a,0} = \lambda_a ~ \forall a \in \mathcal{A}$, batch index $k \leftarrow 0$.

    \For{$t = 1, \dots, T$}{
    ~~Sample $\theta_{a,t} \sim \mathcal{N} \big(\theta_{a}^k,\ \frac{1}{n_a L\gamma } I \big) ~\forall a \in \mathcal{A}$\\
    Choose action $a_t = \argmax_{a \in \mathcal{A}} \alpha_a^\intercal \theta_{a,t}$\\
    Update $k_{a(t)} \leftarrow k_{a(t)} + 1$\\
    \If{$k_{a_t} = 2^{l_{a(t)}}$}{
    $~~l_{a(t)} \leftarrow l_{a(t)} + 1$\\
    Terminate batch $k$ and observe rewards $\{ r_{a_i}\}_{i = B_k}^{t}$ \\
    \For{$ a \in \mathcal{A}$}{
        ~~Update $n_a$ with the number of new samples \\
        Run Algorithm \ref{alg:TS_SGLD} to obtain $\Tilde{\rho}_{a,k+1}$ and $\theta_{a}^{k+1}$ \\
        Update batch index $k \leftarrow k+1$
    }}}

	\caption{Batched Langevin Thompson Sampling (BLTS)}
	\label{alg:TS_Bandit}
\end{algorithm}
\DecMargin{2em}

\vspace{-0.2cm}
\section{Batched Langevin Posterior Sampling For RL}\label{sec:mdp}
In RL frameworks, posterior sampling is commonly used in model-based methods to learn unknown transition dynamics and is known as PSRL\footnote{We also depart from using TS for the RL setting and stick to the more popular posterior sampling terminology for RL.}. In infinite-horizon settings, PSRL operates  by sampling a model and solving for an optimal policy based on the sampled MDP at the beginning of each policy switch. The learner then follows the same policy until the next policy switch. In this context, the concept of a batch corresponds to a policy switch. 

Previous analyses of PSRL have primarily focused on transition distributions that conform to well-behaved conjugate families. Handling transitions that deviate from these families and computing the corresponding posteriors has been heuristically left to MCMC methods. Here, we provably extend PSRL with LMC and introduce \textit{Langevin Posterior Sampling for RL} (LPSRL, Algorithm \ref{alg:TS_alg_MDP_double}) using a static doubling policy-switch scheme. Analyses of PSRL have crucially relied on the true transition dynamics $\theta^*$ and the sampled MDPs being identically distributed \citep{osband2013more, osband2016posterior, russo2014learning, ouyang2017learning, theocharous2017posterior}. However, when the dynamics are sampled from an approximation of the true posterior, this fails to hold. To address the issue, we introduce the Langevin posterior sampling lemma (Lemma \ref{lem:MDP_PS_main_change}), which shows approximate sampling yields an additive error in the Wasserstein-$1$ distance.

\begin{restatable}{lemma}{LangevinPS} \text{(Langevin Posterior Sampling)}\textbf{.}
\label{lem:MDP_PS_main_change}
Let $t_k$ be the beginning time of policy-switch $k$, $\mathcal{H}_{t_k} := \{ s_\tau, a_\tau\}_{\tau=1}^{t_k}$ be the history of observed states and actions till time $t_k$, and $\theta^k \sim \tilde{\rho}_{t_k}$ be the sampled model from the approximate posterior $\tilde{\rho}_{t_k}$ at time $t_k$. Then, for any $\sigma(\mathcal{H}_{t_k})$-measurable function $f$ that is $1$-Lipschitz, it holds that:
\begin{equation}
\label{eqn:mdp_key_prop}
    \Big| \expectation [f(\theta^*) | \mathcal{H}_{t_k}] -  \expectation [f(\theta^k) | \mathcal{H}_{t_k}] \Big| \leq W_1(\tilde{\rho}_{t_k}, \rho_{t_k}).
\end{equation}
By the tower rule, $\Big| \expectation [f(\theta^*) ] - \expectation [f(\theta^{k}) ] \Big| \leq  W_1(\tilde{\rho}_{t_k}, \rho_{t_k})$.
\end{restatable}

As demonstrated later, this error term can be effectively controlled and does not impact the overall regret (Theorems \ref{thm:mdp_general} and \ref{thm:mdp_tabular_main}). It only requires the average reward function $J^{\pi}(\theta)$ to be 1-Lipschitz, as specified in \textbf{Assumption \ref{ass:mdp_lip_J}}\footnote{Mathematical statement is in Appendix \ref{app:assumptions}.}.
Let us consider the parameterization of the transition dynamics $p$ with $\theta \in \reals^d$, where $\theta^* \in \R^d$ denotes the true (unknown) parameter governing the dynamics. We explore two distinct settings based on these parameterizations:

\vspace{-0.2cm}
\begin{itemize}
\setlength\itemsep{0.2em}
    \item \textbf{General Parameterization} (Section \ref{sec:general_mdp}):  In this setting, we consider modeling the full transition dynamics using $\theta^* \in \reals^d$, where $d \ll |\states||\actions|$. This  parameterization can be particularly useful for tackling large-scale MDPs with large (or even continuous) state and action spaces. Towards this end, we consider $\mathcal{S} \cong \R$. Examples of General Parameterization include linear MDPs with feature mappings \cite{jin2020provably}, RL with general function approximation \cite{yang2020function}, and the low-dimensional structures that govern the transition \citep{gopalan15parametrized, yang2020reinforcement}. We provide a real-world example that adopts such parameterization in Appendix \ref{app:mdp_example}.

    Despite \citet{theocharous2017posterior} studies a similar setting, their work confines the parameter space to $\R$. To accommodate a broader class of MDPs, we generalize the parameter space to $\R^d$. As suggested by Theorem \ref{thm:mdp_general}, our algorithm retains the optimal $O(\sqrt{T})$ regret with $O(\log T)$ policy switches, making it applicable to a wide range of general transition dynamics.

    \item \textbf{Simplex Parameterization} (Section \ref{sec:tabular_mdp}): Here, we consider the classical tabular MDPs with finite states and actions. For each state-action pair, there exists a probability simplex $\Delta^{|\mathcal{S}|}$ that encodes the likelihood of transitioning into each state. Hence, in this case, $\theta^* \in \reals^d$ with $d = |\mathcal{S}|^2|\mathcal{A}|$. This structure necessitates sampling transition dynamics from constrained distributions, which naturally leads us to instantiate LPSRL with the \textit{Mirrored Langevin Dynamics} \citep{hsieh2018mirrored} (See Appendix ~\ref{app:MCMC} for more discussions). As proven in Theorem \ref{thm:mdp_tabular_main}, LPSRL with MLD achieves the optimal $O(\sqrt{T})$ regret with $O(\log T)$ policy switches for general transition dynamics subject to the probability simplex constraints.

\end{itemize}

\subsection{The LPSRL Algorithm}

LPSRL (Algorithm \ref{alg:TS_alg_MDP_double}) use \texttt{SamplingAlg} as a subroutine, where SGLD and MLD are invoked respectively depending on the parameterization. Unlike the BLTS algorithm in bandit settings, LPSRL adopts a static doubling batching scheme, in which the decision to move onto the next batch is independent of the dynamic statistics of the algorithm, and thus is algorithmically independent.

Let $t_k$ be the starting time of policy-switch $k$ and let $T_k := 2^{k-1}$ represent the total number of time steps between policy-switch $k$ and $k+1$. At the beginning of each policy-switch $k$, we utilize \texttt{SamplingAlg} to obtain an approximate posterior distribution $\Tilde{\rho}_{t_k}$ and sample dynamics $\theta^k$ from $\Tilde{\rho}_{t_k}$. A policy $\pi_k$ is then computed for $\theta^k$ with any planning algorithm\footnote{We assume the optimality of policies and focus on learning the transitions. When only suboptimal policies are available in our setting, it can be shown that small approximation errors in policies only contribute additive non-leading terms to regret. See details in \cite{ouyang2017learning}.}. The learner follows $\pi_k$ to select actions and transit into new states during the remaining time steps before the next policy switch. New Data is collected all at once at the end of $k$. Once the total number of time steps is being doubled, i.e., $t$ reaches $t_k + T_k -1$, the posterior is updated using the latest data $D$, and the above process is repeated.

\vspace{-0.2cm}
\begin{algorithm}[!ht]
	\setstretch{0.8}
	\SetAlgoLined
	\KwIn{MCMC scheme \texttt{SamplingAlg} initiated with prior $\lambda(\theta)$.}

	\textbf{Initialization}: time step $t \leftarrow 1$, $D \leftarrow \emptyset$ \\
	\For{\text{batch} $k = 1, \dots, K_T$}{
	    $~~T_{k} \leftarrow 2^{k-1} $\\
	    $t_k  \leftarrow 2^{k-1} $ \\
	    Run \texttt{SamplingAlg} and sample $\theta^k$ from posterior: $\theta^k \sim \Tilde{\rho}_{t_k}(\theta | D) $ \\
	   Compute optimal policy $\pi_k$ based on $\theta^k$ \\
	   \For {$t = t_k, t_k + 1, \cdots, t_k + T_k - 1$} {
	   ~~Choose action $a_t \sim \pi_k$ \\
	   Generate immediate reward $\mathcal{ R}(s_t, a_t)$, transit into new state $s_{t+1}$}
    $D \leftarrow D \cup \{ s_t, a_t, \mathcal{R}(s_t, a_t), s_{t+1} \}_{t = t_k}^{t_k + T_k - 1}$
	}
	\caption{Langevin PSRL (LPSRL)}
	\label{alg:TS_alg_MDP_double}
\end{algorithm}

\vspace{-0.2cm}
\subsection{General Parametrization} \label{sec:general_mdp}
In RL context, to study the performance of LPSRL instantiated with SGLD as \texttt{SamplingAlg}, Assumptions \ref{ass:1}-\ref{ass:4} are required to hold on the (unknown) transition dynamics, rather than the (unknown) rewards as in the bandit setting. Additionally, similar to \citet{theocharous2017posterior}, the General Parameterization requires $p(\cdot|\theta)$ to be Lipschitz in $\theta$ (\textbf{Assumption \ref{ass:mdp_lip_transition}}). Mathematical statements of all assumptions are in Appendix \ref{app:assumptions}. We now state the main theorem for LPSRL under the General Parameterization.

\begin{restatable}{theorem}{MDPRegretSGLD}
\label{thm:mdp_general}

Under Assumptions $1-6$, by instantiating \texttt{SamplingAlg} with SGLD and setting the hyperparameters as per Theorem \ref{thm:sgld_convergence}, with $p=2$, the regret of LPSRL (Algorithm \ref{alg:TS_alg_MDP_double}) satisfies:
\begin{align*}
    R_B(T) \leq CH \log T \sqrt{\frac{T}{m}} (d + \log Q + (32 + 8d \kappa^2)p)^{1/2},
\end{align*}
where $C$ is some positive constant, $H$ is the upper bound of the MDP span, and $Q$ denotes the quality of the prior. The total number of iterations required for SGLD is $O(\kappa^2 \log T)$.
\end{restatable}

\paragraph{Discussion.} 
LPSRL with SGLD maintains the same order-optimal regret as exact PSRL in \cite{theocharous2017posterior}. Similar to Theorem \ref{thm:reg_batch_SGLD}, the regret bound has explicit dependence on the quality of prior imposed to transitions, where $\log Q = 0$ when prior is properly centered with its mode at $\theta^*$, or when it is uninformative or flat. Let $\theta^{k,*}$ be the true posterior in policy-switch $k$. Our result relies on $\theta^*$ and $\theta^{k,*}$ being identically distributed, and the convergence of SGLD in $O(\log T)$ iterations to control the additive cumulative error in $\sum_{k=1}^{K_T} T_k W_1 (\Tilde{\rho}_{t_k}, \rho_{t_k})$ arising from approximate sampling.

\subsection{Simplex Parametrization} \label{sec:tabular_mdp}

We now consider the tabular setting where $\theta^*$ specifically models a collection of $|\mathcal{A}|$ transition matrices in $[0, 1]^{|\mathcal{S}| \times |\mathcal{S}|}$. Each row of the transition matrices lies in a probability simplex $\Delta^{|\mathcal{S}|}$, specifying the transition probabilities for each corresponding state-action pair. In particular, if the learner is in state $s \in \mathcal{S}$ and takes action $a \in \mathcal{A}$, then it lands in state $s'$ with $p(s') = p(s' | s, a, \theta^*)$.
In order to run LPSRL on constrained space, we need to sample from probability simplexes and therefore appeal to the Mirrored Langevin Dynamics (MLD) \citep{hsieh2018mirrored} by using the entropic mirror map, which satisfies the requirements set forth by Theorem 2 in \citet{hsieh2018mirrored}.
Under Assumptions \ref{ass:mdp_lip_J} and \ref{ass:mdp_lip_transition}, we have the following convergence guarantee for MLD and regret bound for LPSRL under the Simplex Parameterization.

\begin{restatable}{theorem}{MLDConvergence}
\label{thm:mdp_mld}
At the beginning of each policy-switch $k$, for each state-action pair $(s, a) \in \mathcal{S} \times \mathcal{A}$, sample transition probabilities over $\Delta^{|\mathcal{S}|}$ using MLD with the entropic mirror map. Let $n_{t_k}$ be the number of data samples for any $(s, a)$ at time ${t_k}$, then with step size chosen per \citet{cheng2018convergence}, running MLD with $O(n_{t_k})$ iterations guarantees that
$W_2 (\Tilde{\rho}_{t_k}, \rho_{t_k})  = \tilde{O} \left(  \sqrt{|\mathcal{S}|/n_{t_k}} \right)~.$
\end{restatable}

\begin{restatable}{theorem}{MDPRegretMLD}
\label{thm:mdp_tabular_main}
Suppose Assumptions 5 and 6 are satisfied, then by instantiating \texttt{SamplingAlg} with MLD (Algorithm  \ref{alg:TS_MLD}), there exists some positive constant $C$ such that the regret of LPSRL (Algorithm  \ref{alg:TS_alg_MDP_double}) in the Simplex Parameterization is bounded by 
\[
    R_B(T) \leq C H |\states| \sqrt{|\actions| T \log(|\states||\actions|T) },
\]
where $C$ is some positive constant, $H$ is the upper bound of the MDP span. The total number of iterations required for MLD is $O(|\mathcal{S}|^2|\mathcal{A}|^2 T)$.
\end{restatable}

\paragraph{Discussion.} In simplex parameterization, instantiating LPSRL with MLD achieves the same order-optimal regret, but the computational complexity in terms of iterations for MLD is linear in $T$ as opposed to $\log T$ for SGLD in the General Parameterization. Nevertheless, given that the simplex parameterization implies simpler structures, we naturally have fewer assumptions for the theory to hold.

\vspace{-0.2cm}
\section{Experiments}
\label{sec:empirical}
In this section, we perform empirical studies in simulated environments for bandit and RL to corroborate our theoretical findings. By comparing the actual regret (average rewards) and the number of batches for interaction (maximum policy switches), we show \textit{Langevin TS} algorithms empowered by LMC methods achieve appealing statistical accuracy with low communication cost. For additional experimental details, please refer to Appendix \ref{sec:appendix_empirical}.

\subsection{Langevin TS in Bandits}
We first study how Langevin TS behaves in learning the true reward distributions of log-concave bandits with different priors and batching schemes. Specifically, we construct two bandit environments\footnote{Our theories apply to bandits with a more general family of reward distributions.} with Gaussian and Laplace reward distributions, respectively. While both environments are instances of log-concave families, Laplace bandits do not belong to conjugate families.

\subsubsection{Gaussian Bandits}
We simulate a Gaussian bandit environment with $N=15$ arms. The existence of closed-form posteriors in Gaussian bandits allows us to benchmark against existing exact TS algorithms. More specifically, we instantiate Langevin TS with SGLD (SGLD-TS), and perform the following tasks:
\vspace{-0.5em}
\begin{itemize}
    \setlength\itemsep{0.1em}
    \item Compare SGLD-TS against both frequentist and Bayesian methods, including UCB1, Bayes-UCB, decaying $\epsilon$-greedy, and exact TS.
    \item Apply informative priors and uninformative priors for Bayesian methods based on the availability of prior knowledge in reward distributions.
    \item Examine all methods under three batching schemes: fully-sequential mode, dynamic batch, static batch.
\end{itemize}



\begin{figure*}[!ht]
    \centering
    \begin{subfigure}{0.32\linewidth}
        \centering
        \includegraphics[width=\linewidth]{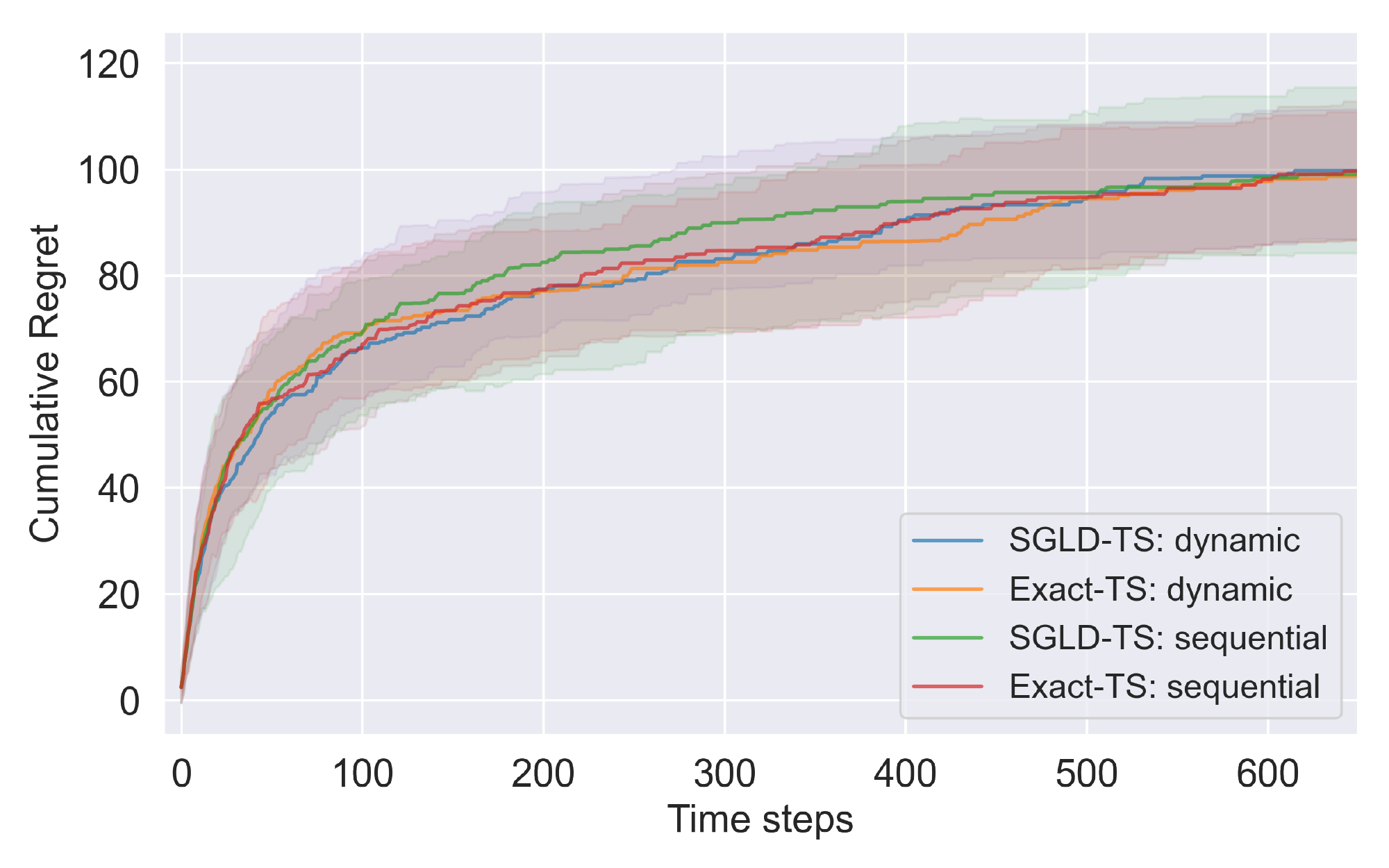}
    \end{subfigure}
    \begin{subfigure}{0.32\linewidth}
        \centering
        \includegraphics[width=\linewidth]{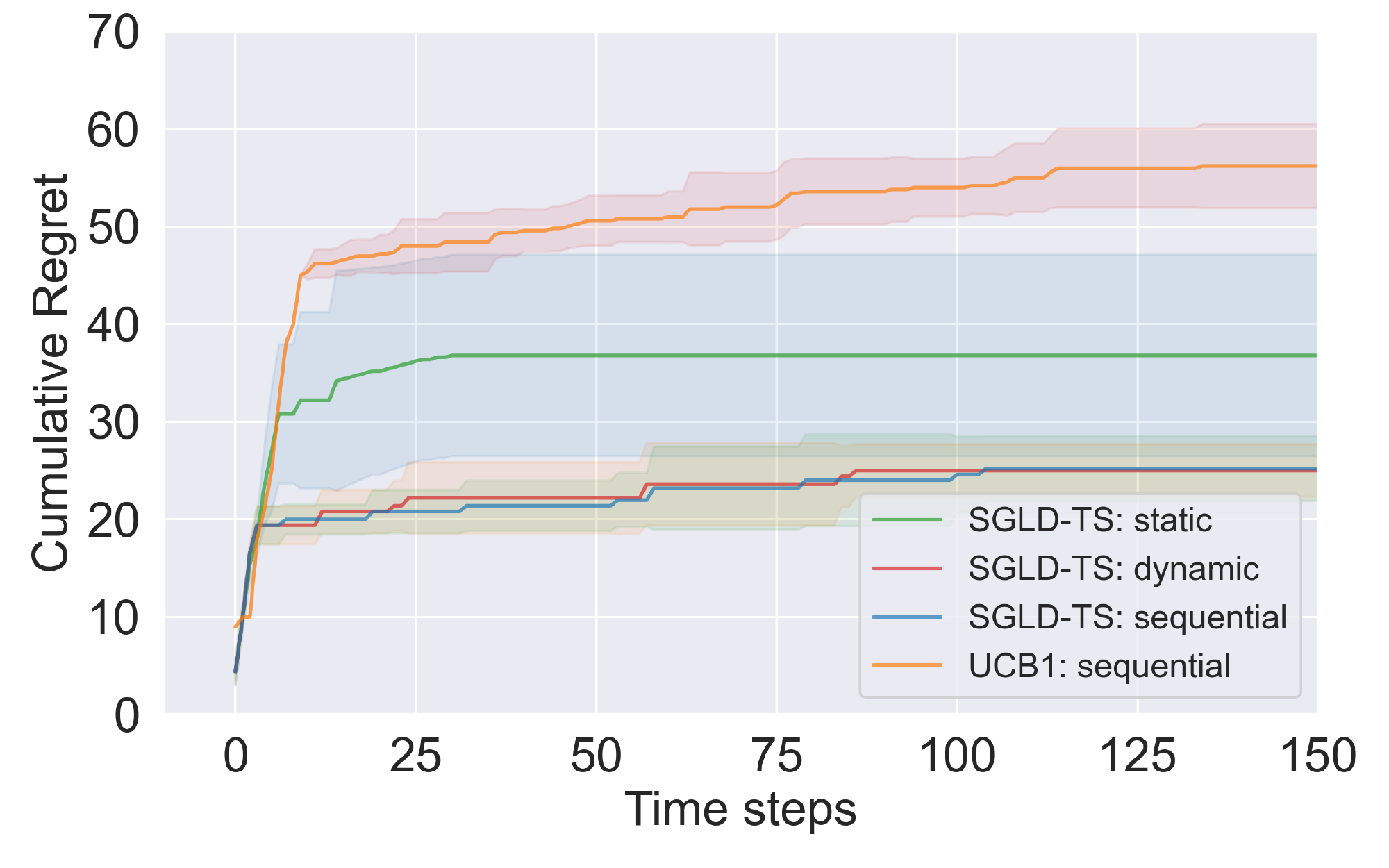}
    \end{subfigure}
    \begin{subfigure}{0.32\linewidth}
        \centering
        \includegraphics[width=\linewidth]{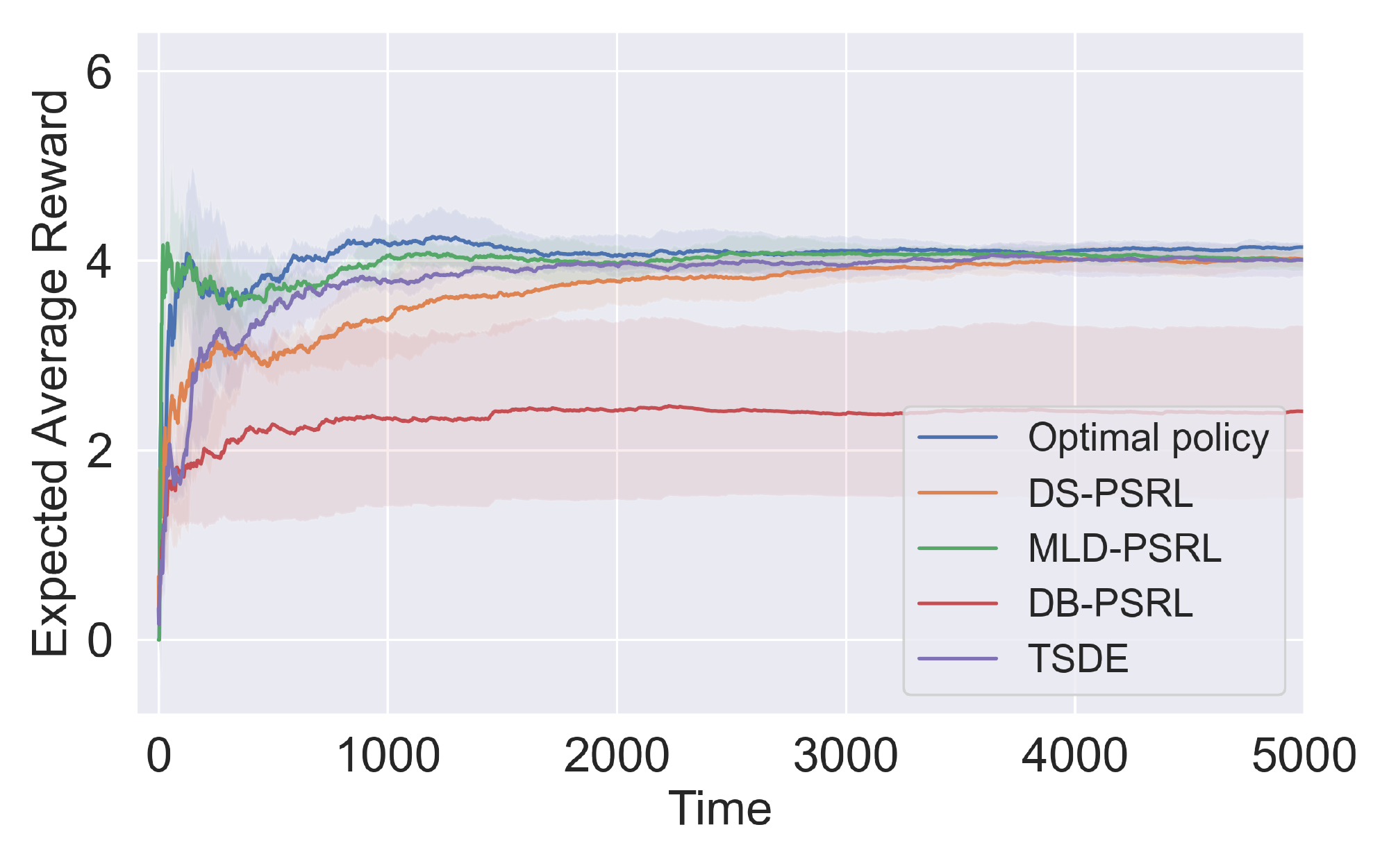}
    \end{subfigure}
    \caption{(a)~Regret in Gaussian Bandits ($N = 15$): expected regret is reported over 10 experiments with informative priors. Results show SGLD-TS under dynamic batching scheme achieves optimal performance as in the sequential case without using approximate sampling. Results with uninformative priors yield the same conclusions (See Appendix \ref{sec:appendix_empirical}). (b)~Regret in Laplace Bandits ($N = 10$): regret is reported over 10 experiments with informative priors. As in Gaussian Bandits, SGLD-TS with dynamic batching scheme achieves optimal regret and outperforms UCB1. (c)~Average reward in RiverSwim: expected average reward is reported over 10 experiments. MLD-PSRL achieves optimal average reward upon convergence with a small number of policy switches.}
    \label{fig:regret}
\end{figure*}

\textbf{Results and Discussion.} Figure~\ref{fig:regret}(a) illustrates the cumulative regret for SGLD-TS and Exact-TS with favorable priors. Table~\ref{tab:exp_bandit_reg} reports the regret upon convergence along with the total number of batches in interaction. Note that SGLD-TS equipped with dynamic batching scheme implements Algorithm~\ref{alg:TS_Bandit} (BLTS).  
Empirical results demonstrate that SGLD-TS is comparable to Exact-TS under all batching schemes, and is empirically more appealing compared to UCB1 as well as Bayes-UCB. While static batch incurs slightly lower communication costs compared to dynamic batch, results show that all methods under dynamic batch scheme are more robust with smaller standard deviation. Our BLTS algorithm thus well balances the trade-off between statistical performance, communication, and computational efficiency by achieving the order-optimal regret with a small number of batches.

\begin{table}[t]
    \def\arraystretch{1.5}
    \centering
    
    \small
    \begin{tabular}{|>{\centering\arraybackslash}m{3cm}
               |>{\centering\arraybackslash}m{2cm}
               |>{\centering\arraybackslash}m{2cm}
               |>{\centering\arraybackslash}m{2cm}
                |>{\centering\arraybackslash}m{2cm}|
                 |>{\centering\arraybackslash}m{2cm}|}
\hline 
  & \textbf{SGLD-TS} & \textbf{Exact-TS} & \textbf{UCB1} & \textbf{Bayes-UCB} & \textbf{Batches} \\
 \hline
 \textbf{Fully sequential} & $99.66 \pm 13.09$ & $99.07 \pm 12.23$ & $154.13 +- 4.10$ & $160.55 \pm 25.75$ & $650.0 \pm 0.0 $ \\
 \cline{1-6}
 \textbf{Static batch} & $148.52 \pm 39.28$ & $145.94 \pm 31.46$ & $155.17 \pm 5.06$  & $231.80 \pm 52.11$ & $9.0 \pm 0.0 $\\
 \cline{1-6}
 \textbf{Dynamic batch} & $99.80 \pm 15.62$ & $98.71 \pm 12.10$ & $153.31 \pm 3.83$ & $214.43 \pm 0.5$ & $22.93 \pm 1.50 $\\
 \hline
    \end{tabular}
    \vspace{-0.2cm}
    \caption{Average regret with the standard deviation under different batching schemes. The last column quantifies communication cost $w.r.t$ the total number of batches for interaction. BLTS (SGLD-TS under dynamic batching scheme) achieves order-optimal regret with low communication cost.} 
    \label{tab:exp_bandit_reg}
\end{table}

\subsubsection{Laplace Bandits}
To demonstrate the applicability of Langevin TS in scenarios where posteriors are intractable, we construct a Laplace bandit environment with $N = 10$ arms. It is important to note that Laplace reward distributions do not have conjugate priors, rendering exact TS inapplicable in this setting. Therefore, we  compare the performance of SGLD-TS with favorable priors against UCB1. Results presented in Figure \ref{fig:regret}(b) reveal that, similar to the Gaussian bandits, SGLD-TS with dynamic batching scheme achieves comparable performance as in the fully-sequential setting and significantly outperforms UCB1, highlighting its capability to handle diverse environments. In addition, the static batching scheme exhibits larger deviations compared to the dynamic batching, which aligns with results in Table \ref{tab:exp_bandit_reg}.

\subsection{Langevin PSRL in Average-reward MDPs}
In MDP setting, we consider a variant of RiverSwim environment \citep{strehl2008analysis}, which is a common testbed for provable RL methods. Specifically, it models an agent swimming in the river with five states, two actions ($|\mathcal{S}| = 5, |\mathcal{A}| = 2$). In this tabular case, LPSRL (Algorithm \ref{alg:TS_alg_MDP_double}) employs MLD (Algorithm \ref{alg:TS_MLD} in Appendix \ref{app:MCMC}) as \texttt{SamplingAlg}, namely, MLD-PSRL. We benchmark the performance of MLD-PSRL against other mainstream model-based RL methods, including TSDE \citep{ouyang2017learning}, DS-PSRL \citep{theocharous2017posterior} and DB-PSRL (exact-PSRL\cite{strens2000bayesian} with dynamic batch). Note that MLD-PSRL and DS-PSRL adopt the static doubling policy switch scheme discussed in section \ref{sec:mdp}. Dynamic doubling policy switch scheme adopted by both DB-PSRL and TSDE is akin to the one we use in bandit setting, but based on the visiting counts of state-action pairs. We simulate $10$ different runs of experiment, and report the average rewards obtained by each method in Figure \ref{fig:regret}(c). Mechanisms used by each method are summarized in Table~\ref{tab:exp_mdp_avg_reward}, along with the average rewards achieved and maximum number of policy switches incurred. 


 \begin{table}[!ht]
    \def\arraystretch{1.2}
    \centering
    
    \small
    \begin{tabular}{|>{\centering\arraybackslash}m{3.0cm}
               |>{\centering\arraybackslash}m{2cm}
               |>{\centering\arraybackslash}m{2cm}
               |>{\centering\arraybackslash}m{2cm}
               |>{\centering\arraybackslash}m{2cm}
                |>{\centering\arraybackslash}m{2.8cm}|}
\hline 
  & \textbf{MLD-PSRL} & \textbf{DS-PSRL} & \textbf{DB-PSRL} & \textbf{TSDE} & \textbf{Optimal policy} \\
 \cline{1-6}
 Static ps & $\surd$ & $\surd$ &  & &  \\
 \cline{1-6}
 Dynamic ps & &  & $\surd$ & $\surd$ & \\
 \cline{1-6}
 Linear growth & &  & & $\surd$ & \\
 \cline{1-6}
 \textbf{Avg. reward} & $4.01 \pm 0.11$ & $4.02 \pm 0.08$ & $2.41 \pm 0.91$ & $4.01 \pm 0.17$ & $4.15 \pm  0.04$ \\
 \cline{1-6}
 \textbf{Max. switches} & $12.0 \pm 0.0$ & $12.0 \pm 0.0$ & $15.33 \pm 1.70$ & $94.0 \pm 3.56$ & - \\
 \hline
    \end{tabular}
    \vspace{-0.2cm}
    \caption{We report the average reward and the maximum number of policy switches all methods over  10 different runs. MLD-PSRL instantiates Alogrithm~\ref{alg:TS_alg_MDP_double} in Section~\ref{sec:mdp}, which achieves order-optimal performance with small number of policy switches.} 
    \label{tab:exp_mdp_avg_reward}
\end{table}

\textbf{Results and Discussion.} We demonstrate that MLD-PSRL achieves comparable performance compared to existing PSRL methods while significantly reducing communication costs through the use of static policy switches. In contrast, as illustrated in Figure \ref{fig:MDP_policy_switches} (Appendix~\ref{sec:appendix_empirical}) and Table~\ref{tab:exp_mdp_avg_reward}, TSDE achieves near-optimal performance but requires high communication costs. Additionally, our empirical results reveal that the static policy switch in the MDP setting outperforms the dynamic policy switch alone. This observation aligns with existing findings that frequent policy switches in MDPs can harm performance. Moreover, compared to DS-PSRL, MLD-PSRL is applicable to more general frameworks when closed-form posterior distributions are not available\footnote{Different mirror maps can be applied to MLD method depending on parameterization.}.

\vspace{-0.2cm}
\section{Conclusion} 


In this paper, we jointly address two challenges in the design and analysis of Thompson sampling (TS) methods. 
Firstly, when dealing with posteriors that do not belong to conjugate families, it is necessary to generate approximate samples within a reasonable computational budget. Secondly, when interacting with the environment in a batched manner, it is important to limit the amount of communication required. 
These challenges are critical in real-world deployments of TS, as closed-form posteriors and fully-sequential interactions are rare. In stochastic MABs, approximate TS and batched interactions are studied independently. We bridge the two lines of work  by providing a Langevin TS algorithm that works for a wide class of reward distributions with only logarithmic communication. In the case of undiscounted infinite-horizon MDP settings, to the best of our knowledge, we are the first to provably incorporate approximate sampling with the TS paradigm. This enhances the applicability of TS for RL problems with low communication costs. Finally, we conclude with experiments to demonstrate the appealing empirical performance of the Langevin TS algorithms.

\section*{Acknowledgements}
This work is supported in part by the National Science Foundation Grants NSF-SCALE MoDL(2134209) and NSF-CCF-2112665 (TILOS), the U.S. Department Of Energy, Office of Science, and the Facebook Research award. Amin Karbasi acknowledges funding in direct support of this work from NSF (IIS-1845032), ONR (N00014- 19-1-2406), and the AI Institute for Learning-Enabled Optimization at Scale (TILOS).

\clearpage
\bibliographystyle{apalike}
\bibliography{ts_main}

\clearpage
\onecolumn
\appendix
\noindent\textbf{\Large{Appendices}}

\section{MCMC Methods}
\label{app:MCMC}

\subsection{Unconstrained Approximate Sampling}
Suppose a target distribution $\rho$ is parameterized by $\theta \in \R^d$, and observed data $\{X_i\}_{i=1}^n$ are independently identically distributed. A posterior distribution defined up to a normalization factor can be expressed via the Gibbs distribution form:
\[
   \rho ( \theta | X_1, \dots, X_n) \propto \lambda(\theta) \prod_{i=1}^n p(X_i; \theta) = \exp\lrp{-U(\theta)},
\]
where $\lambda(\theta)$ is the prior distribution of $\theta$, $p(X_i; \theta)$ is the likelihood function, and $U(\theta) := -\log \lrp{ \lambda(\theta) } - \sum_{i=1}^n \log \lrp{ p(X_{i}; \theta)}$ is the energy function.

Typical MCMC methods require computations over the whole dataset, which is inefficient in large-scale online learning. To overcome this issue, we adopt SGLD \cite{welling2011bayesian} as one of the approximate sampling methods, which is developed upon stochastic optimization over mini-batch data $D \subseteq \{X_i\}_{i=1}^n$. The update rule is based on the Euler-Murayama discretization of the Langevin \textit{stochastic differential equation (SDE)}:
\begin{align*}
   d \theta_{ t} = \frac{1}{2} \lrp{ \nabla \log \lrp{ \lambda(\theta_{0}) } +  \frac{n}{|D|} \sum_{i \in D} \nabla \log \lrp{p(x_i; \theta_{t})} } dt + \sqrt{2} d B_t,
\end{align*}
where $B_t$ is a Brownian motion. To further improve computation, we reuse samples from previous batches to warm start the Markov chains (Algorithm \ref{alg:TS_SGLD}). The resulting dependent structure in samples will complicate our analysis. 

\subsection{Constrained Approximate Sampling}
While the convergence of SGLD methods is well-studied, it is only applicable to unconstrained settings. 
To enable sampling from constrained non log-concave distributions, such as probability simplex in transition dynamics of MDPs, reparameterization can be used in conjunction with SGLD. Alternatively, one can adopt MLD \cite{hsieh2018mirrored} which utilizes mirror maps for sampling from a dual unconstrained space (Algorithm \ref{alg:TS_MLD}). Let the probability measure of $\theta$ be $d \rho = e^{-U(\theta)} \text{d} \theta$, where dom($U$) is constrained. Suppose there exist a mirror map $h$ that maps $\rho$ to some unconstrained distribution $d \nu = e^{-W(\omega)} \text{d} \omega$, denoted by $\nabla h \#\rho = \nu$. Then MLD has the following SDE:
\begin{equation}
\label{eqn:mld}
    \begin{cases}
    \text{d} \omega_t = - (\nabla W \circ \nabla h)(\theta_t) \text{d} t +\sqrt{2} \text{d} B_t \\
    \theta_t = \nabla h^*(\omega_t)
    \end{cases}
    ,
\end{equation}
where $h^*$ is the dual of $h$, and $(\nabla h)^{-1} = \nabla h^*$.

\begin{algorithm}[ht]
	\setstretch{0.8}
	\SetAlgoLined
	\KwIn{$\mathcal{S}, \mathcal{A}$, mirror map $h$, observed transitions $\{X_s\}_{s=1}^n$, total iterations $N$}

	\For{$i = 1, \dots, N$}{
	            Subsample $D \subseteq \{X_s\}_{s=1}^n$ \\
	            Sample $\omega_{i+1} \sim \nabla h \#e^{-U}$ from the unconstrained dual space \\
	            Compute constrained sample $\theta_{i+1} = \nabla h^*(\omega_{i+1})$
	        }
	       \KwOut{ $\theta_N$}
	\caption{Mirrored Langevin Dynamics (MLD)}
	\label{alg:TS_MLD}
\end{algorithm}
\DecMargin{1em}

In tabular settings of MDP, MLD needs to be run against each row of the $|\mathcal{A}| \times |\mathcal{S}|$ matrices to generate a sampled transition from simplex $\Delta_{|\states|}$ for each state-action pair. In this case, entropic mirror map will be adopted as $h$, which is given by \begin{equation} 
\label{eqn:mld_entropic}
    h(\theta) = \sum_{i=1}^{|\states|} \theta_i \log \theta_i + (1 - \sum_{i=1}^{|\states|} \theta_i) \log (1 - \sum_{i=1}^{|\states|} \theta_i), ~~~~\text{where}~~ 0 \log 0 := 0.
\end{equation}

\section{Assumptions}\label{app:assumptions}

Here we explicitly mention all of the assumptions required in the paper. Assumptions 1-4 are required for SGLD to converge (Algorithm \ref{alg:TS_SGLD} and Theorem \ref{thm:sgld_convergence}), Assumptions \ref{ass:mdp_lip_J} and \ref{ass:mdp_lip_transition} are required in Section \ref{sec:mdp}.

\begin{assumption} [Assumption on the family $p(S|\theta)$ for approximate sampling] 
\label{ass:1}
Assume that $\log p(s|\theta)$ is $L$-smooth and $m$-strongly concave over $\theta$:
\begin{align*}
    &- \log p (s|\theta') - \nabla_\theta \log p (s|\theta')^\top (\theta - \theta') + \frac{m}{2} \lrn{\theta - \theta' }^2 \leq - \log p (s|\theta)\\
    &\leq - \log p (s|\theta') - \nabla_\theta \log p (s|\theta')^\top (\theta - \theta') + \frac{L}{2}\lrn{\theta - \theta'}^2 ~~ \forall \theta, \theta' \in \reals^{d}, s \in \mathcal{S}
\end{align*}   
\end{assumption}

\begin{assumption} [Assumption on true reward/transition distribution $p(S|\theta^*)$]
Assume that $p(S; \theta^*)$ is strongly log-concave in $S$ with some parameter $\nu$, and that $\nabla_\theta \log p(s|\theta^*)$ is $L$-Lipschitz in $S$:
\[
    -(\nabla_s \log p(s | \theta^*) - \nabla_s \log p (s' | \theta^*))^\top (s - s') \geq \nu \lrn{s-s'}^2, ~~ \forall s,s' \in \reals
\]
\[
    \lrn{\nabla_\theta \log p(s|\theta^*) - \nabla_\theta \log p (s'|\theta^*)} \leq L \lrn{s-s'}, ~~ \forall s, s' \in \reals
\]
    
\end{assumption}

\begin{assumption}[Assumption on the prior distribution]
Assume that $\log \lambda (\theta)$ is concave with $L$-Lipschitz gradients for all $\theta \in \reals^d$:
\[
    \lrn{\nabla_\theta \lambda (\theta) - \nabla_\theta \lambda (\theta')} \leq L \lrn{\theta - \theta'}, ~~ \forall \theta, \theta' \in \reals^d
\]
\end{assumption}

\begin{assumption} [Joint Lipschitz smoothness of $\log p(S|\theta)$]
\label{ass:4}
\[
    \lrn{ \nabla_\theta \log p(s|\theta) - \nabla_\theta \log p(s'|\theta)} \leq L \lrn{\theta - \theta'} + L \lrn{s - s'}, ~~ \forall \theta, \theta' \in \reals^d, s, s' \in \reals
\]
    
\end{assumption}

\begin{assumption} [1- Lipschitzness of $J(\theta)$ in $\theta$]
\label{ass:mdp_lip_J}
The optimal average-reward
function $J$ satisfies
\[
    \lrn{J(\theta) - J(\theta^\prime)} \leq  \|\theta - \theta^\prime\|, ~~ \forall \theta, \theta^\prime \in \reals^d
\]
where $J(\theta) = \max_{\pi} J^{\pi}(\theta)$.

\end{assumption} 

\begin{assumption} [Lipschitzness of transition in $\theta$ for RL] 
\label{ass:mdp_lip_transition}
There exists a constant $L_p$ such that the transition for each state-action pair is $L_p$-Lipschtiz in parameter space:

\[
    \lrn{p(\cdot| s, a, \theta) - p(\cdot |  s, a, \theta^\prime)} \leq L_p \lrn{\theta - \theta^\prime}, ~~ \forall \theta, \theta' \in \reals^d,~~s, a \in \mathcal{S} \times \mathcal{A}
\]
    
\end{assumption}

\section{Convergence of SGLD with Batched Data}
\label{sec:appendix_sgld}

\allowdisplaybreaks
\counterwithin{lemma}{section}

In this section, we prove the convergence of SGLD in sequential decision making frameworks under the batch scheme, which is stated with precise hyperparameters as Theorem \ref{thm:sgld_convergence_app}. We first state the supporting lemmas, followed by the proof of the convergence theorem. 

\begin{lemma} [Lemma 5 in \cite{mazumdar2020thompson}]
\label{lem:original_lemma_5}
Denote $\widehat{U}$ as the stochastic estimator of $U$.
Then for stochastic gradient estimate with $k$ data points, we have, 
\begin{align*}
    \E{\lrn{ \nabla \widehat{U}(\theta) - \nabla U(\theta) }^p \big| \theta} \leq 2 \frac{n^{p/2}  }{k^{p/2}} \lrp{ \frac{\sqrt{d p} L_a^* }{\sqrt{\nu_a}} }^p.
\end{align*}
\end{lemma}

\begin{lemma}[Lemma 6 from \cite{mazumdar2020thompson}]
\label{lem:sgld_batch_convergence}
For a fixed arm $a$ with $n$ samples, suppose we run Algorithm \ref{alg:TS_SGLD}  with step size $\eta \leq \frac{\hat{m}}{32 \hat{L}^2}$ for N iterations to generate samples from posterior $\rho_n^* \propto \exp (-U)$, in which $U$ is $\hat{m}-$strongly convex and $\hat{L}-$Lipschitz smooth. If at each step $i \in [N]$, the $p$-th moment between the true gradient and the stochastic gradient satisfies
$
    \E{\| \nabla U (\theta_{i\eta}) - \nabla \hat{U} (\theta_{i\eta})\|^p ~|~\theta_{i\eta}  } \leq \Delta_p, 
$
then:
\[
W_p^p (\tilde{\rho}_{i\eta, n}, \rho_n^*) \leq \left(1 - \frac{\hat{m}}{8}\eta\right)^{pi}  W_p^p (\rho_0, \rho_{n}^*) + 2^{5p}\frac{\hat{L}^p}{\hat{m}^{p}}(dp)^{p/2} (\eta)^{p/2} + 2^{2p+3}\frac{\Delta_p}{\hat{m}^p}
\]

where $\rho_0 = \tilde{\rho}_{0\eta, n}$.

\end{lemma}

\begin{theorem} [SGLD convergence]
\label{thm:sgld_convergence_app}
Fix an arm $a \in \mathcal{A}$ and suppose that Assumptions 1-4 are met for it.  Let $\kappa := \max \{L / m, L / \nu\}$, $n_{k}$ be the number of available rewards for arm $a$ when running SGLD for the $k$-th time, $\rho_{a, n_{k}}$ be the exact posterior of arm $a$ after observing $n_{k}$ samples, and $\tilde{\rho}_{a, n_{k}}$ be the corresponding approximate posterior obtained by SGLD. If $\mathbb{E}_{\theta \sim \rho_{a, n_{k}}} [ \|\theta - \theta^*\|^p ]^{1/p} \leq \frac{\Tilde{D}}{\sqrt{n_{ k}}} $ is satisfied by the posterior, then with mini-batch size $s = \frac{32L^2}{m \nu} = \mathcal{O}(\kappa^2)$, step size $\eta = \frac{mn_{k}}{32 L^2 (n_{k} + 1)^2} = \mathcal{O}(\frac{1}{L \kappa n_{k}})$, and the number of steps $N =  \frac{1280 L^2 (n_{k} + 1)^2}{m^2 n_{k}^2} = \mathcal{O}(\kappa^2)$, SGLD in Algorithm \ref{alg:TS_SGLD} converges in Wasserstein-$p$ distance: 
\[
    W_p \left(\tilde{\rho}_{a, n_{k}}, \rho_{a, n_{k}}\right) \leq \frac{2 \Tilde{D}}{\sqrt{n_k}}, ~~~~ \forall \Tilde{D} \geq \sqrt{\frac{32d p}{m}},~ p \geq 2.
\]
\end{theorem}

\begin{proof} [Proof of Theorem ~\ref{thm:sgld_convergence_app}]
The proof follows similarly to that of Theorem 6 in \cite{mazumdar2020thompson}. Compared to the analysis in \cite{mazumdar2020thompson}, our proof is based on induction on the batches, as opposed to induction on the number of samples, as for us, SGLD is only  executed at the end of the batch. Let $B_k$ be the $k$-th batch. Now for the base case, i.e. when $k = 1$, we have that $n_k=1$. And therefore the claim follows by the initialization of the algorithm (this is similar to the fully sequential case in \cite{mazumdar2020thompson}).

Now, suppose that the claim holds for batch $k-1$. That is, suppose that all the necessary conditions are met and that $W_p \left(\tilde{\rho}_{a,n_{k-1}}, \rho_{a,n_{k-1}}\right) \leq \frac{2\Tilde{D}}{\sqrt{n_{k-1}}}$. 

Taking the initial condition $\rho_0 = \tilde{\rho}_{a, n_{k-1}}$ in Lemma \ref{lem:sgld_batch_convergence}, we get that:

\[
    W_p^p (\tilde{\rho}_{i\eta, n_k}, \rho_{n_k}^*) \leq \left(1 - \frac{\hat{m}}{8}\eta\right)^{pi}  W_p^p (\tilde{\rho}_{a, n_{k-1}}, \rho_{n_k}^*) + 2^{5p}\frac{\hat{L}^p}{\hat{m}^{p}}(dp)^{p/2} (\eta)^{p/2} + 2^{2p+3}\frac{\Delta_p}{\hat{m}^p}.
\]

Now we know that:

\begin{align*}
    W_p (\rho_{n_k}^*, \tilde{\rho}_{a, n_{k-1}}) &\leq W_p (\rho_{n_k}^*, \rho_{n_{k-1}}^*) + W_p (\rho_{n_{k-1}}^*, \tilde{\rho}_{a, n_{k-1}})\\
    &\leq \frac{\Tilde{D}}{\sqrt{n_{ k}}} + \frac{\Tilde{D}}{\sqrt{n_{ k-1}}} + \frac{2\Tilde{D}}{\sqrt{n_{ k-1}}}\\
    &\leq \frac{8\Tilde{D}}{\sqrt{n_{ k}}}
\end{align*}

where the first inequality follows from triangle inequality, the second one follows from the assumption on the posterior and the induction hypothesis, and the last one just upper bounds the expression while also using the fact that $n_k \leq 2 n_{k-1}$. This shows us that we can get the same upper bound as is seen in the fully sequential proof. The main point to note is that the proof has enough looseness in it, so that despite collecting at most double the data, the same bounds hold. With the choice of hyperparameters, taking $i=N$ and using Lemma \ref{lem:original_lemma_5} leads us to the conclusion that $W_p \left(\tilde{\rho}_{a, n_{k}}, \rho_{a, n_{k}}\right) \leq \frac{2 \Tilde{D}}{\sqrt{n_k}}$~.

\end{proof}

We now state the concentration results provided by SGLD in Lemma~\ref{lem:sgld_concentration}, which shows the probability that the sampled parameters (from the approximate posterior) are far away from the true dynamics is small.  Lemma~\ref{lem:sgld_concentration} extends Lemma 11 in \cite{mazumdar2020thompson} to the batched settings.

\begin{lemma} [Concentration of SGLD in bandits]
\label{lem:sgld_concentration} 
For a fixed arm $a \in \mathcal{A}$, say that it is pulled $n_{k-1}$ times till batch $k-1$ and $n_k$ times till batch $k$ (where $n_k \leq 2 n_{k-1}$).
 Suppose that Assumptions \ref{ass:1}-\ref{ass:4} are satisfied, then for $\delta \in (0, 1)$, with parameters as specified in Theorem~\ref{thm:sgld_convergence_app}, the sampled parameter $\theta_{a}^k$ generated in the $k$-th batch satisfies,
\[
    \mathbb{P}_{\theta_{a}^k \sim \tilde{\rho}_{a,n_k} [\gamma]} \left( \| \theta_{a}^k - \theta_a^* \|_2 > \sqrt{\frac{36e}{n_k m} \left(d + \log Q_a + 2 \sigma \log 1/\delta + 2(\sigma + \frac{m d}{18 L \gamma})\log 1/\delta \right)} ~~ \Bigg| ~ Z_{k-1} \right) < \delta,
\] 

where $Z_{k-1} = \{ \lrn{\theta_{a}^{k-1} - \theta_a^*}_2 \leq C(n_k)$ \}, $C(n_k) = \sqrt{\frac{18e}{n_k m}} (d + \log Q_a + 2 \sigma \log 1/\delta)^{0.5}$, $\sigma = 16 + \frac{4dL^2}{\nu m}$.

\end{lemma}

\begin{proof} [Proof of Lemma \ref{lem:sgld_concentration}]

The proof follows exactly as Lemma 11 from \cite{mazumdar2020thompson} by replacing the notations in fully-sequential settings by those in batched settings, i.e., $\theta_{a, t}$ by $\theta_a^k$, $\theta_{a, t-1}$ by $\theta_a^{k-1}$.

\end{proof}

\section{Proofs of Langevin Thompson Sampling in Multi-armed Bandits}
\label{app:bandit}

\allowdisplaybreaks
\counterwithin{lemma}{section}

In this section, we provide the regret proofs of BLTS algorithm in the stochastic multi-armed bandit (MAB) setting, which are discussed in Section~ \ref{sec:bandits_main}. In particular, we discuss the information exchange guarantees under dynamic batching scheme and its communication cost. 
We then utilize the convergence of SGLD in Appendix \ref{sec:appendix_sgld} and the above results to prove the problem-dependent regret bound in MAB setting.

\subsection{Notations}
We first introduce the notation being used in this section, which is summarized in Table \ref{tab:notation}.
\begin{table*}[!ht]
    \def\arraystretch{1.5}
    \centering
    \small
    \begin{tabular}{|>{\centering\arraybackslash}m{2cm}
               |>{\centering\arraybackslash}m{12cm}|}
\hline 
  \textbf{Symbol} & \textbf{Meaning} \\
 \hline
  $\mathcal{A}$ & set of arms in bandit environment  \\
 \hline
 $N$ & number of arms in bandit environment, i.e., $|\mathcal{A}|$   \\
 \hline
 $T$ & time horizon   \\
 \hline
 $K$ & total number of batches   \\
 \hline
 $B(t)$ & starting time of the batch containing timestep $t$  \\
 \hline
 $B_k$ & starting time of the $k$-th batch  \\
 \hline
 $l_a$ & trigger of dynamic batches (a batch is formed when $k_a(t) = 2^{l_a}$), a monotonically-increasing integer for arm $a$  \\
 \hline
 $k_a(t)$ & the number of times that arm $a$ has been pulled up to time $t$  \\
 \hline
 $p_a(r| \theta_a)$ & reward distribution of arm $a$ parameterized by $\theta_a \in \R^d$   \\
 \hline
 $\theta_a$ & parameter of reward distribution for arm $a \in \mathcal{A}$  \\
 \hline
 $\mu_a$ & expected reward of arm $a$, $\mu_a := \mathbb{E}[r_a | \theta_a^*]$ \\
 \hline
 $\hat{\mu}_a$ & estimated expected reward of arm $a$, $\hat{\mu}_a := \mathbb{E}[\lrn{\theta_a}]$ \\
 \hline
  $Q_a$ & quality of prior for arm $a$, $Q_a := \max_{\theta}\frac{ p_a(\theta)}{p_a(\theta_a^*)}$  \\
 \hline
 $\kappa$ & condition number of parameterized reward distribution, $\kappa := \max \{L / m, L / \nu\}$
  \\
 \hline
 $\lambda_a(\theta_a)$ & prior distribution over $\theta_a \in \R^d$   \\
 \hline
 $U$ & energy function of posterior distribution $\rho$ : $\rho \propto e^{-U}$   \\
 \hline
 $L$ & Lipschitz constant of the true reward distribution and likelihood families $p_a(r|\theta^*)$ in $r$  \\
 \hline
 $m$ & strong log-concavity parameter of $p_a(r; \theta)$ in $\theta$ for all $r$  \\
 \hline
 $\nu$ & strong log-concavity parameter of $p_a(r; \theta)$ in $r$  \\
 \hline
    \end{tabular}
    \caption{Notations in multi-armed bandit setting.} 
    \label{tab:notation}
\end{table*}

\subsection{Communication cost of Dynamic Doubling Batching Scheme}
\label{sec:appendix_mab_batch}

In batched setting, striking a balance between batch size and the number of batches is critical to achieving optimal performance. More specifically, it is crucial to balance the number of actions taken within each batch, with the frequency of starting new batches to collect new data and update the posteriors. According to Lemma~\ref{lem:bandit_pull_relation}, dynamic doubling batching scheme guarantees an arm  that has been pulled $k$ times has at least $k/2$ observed rewards, indicating that communication between the learner and the environment is sufficient under this batching scheme.

\begin{restatable}{lemma}{banditPull}
\label{lem:bandit_pull_relation}
    Let $t$ be the current time step, $B(t)$ be the starting time of the current batch, $k_a(t)$ be the number of times that arm $a$ has been pulled up to time $t$. For all $a \in \mathcal{A}$, the dynamic batch scheme  ensures: 
    \[\frac{1}{2} k_a (t) \leq k_a (B(t)) \leq k_a (t).\]
\end{restatable}

\begin{proof}[Proof of Lemma~\ref{lem:bandit_pull_relation}]
    By the mechanism of our batch scheme, a new batch will begin when the number of times of any arm $a \in \mathcal{A}$ being pulled is doubled. It implies that the number of times that an arm is pulled within a batch is less than the number of times that it has been pulled at the beginning of this batch. At any time step $t \leq T$:
    \[
    k_a(t) - k_a(B(t)) \leq k_a(B(t)),
    \]
    which gives $\frac{1}{2}k_a(t) \leq k_a(B(t))$. On the other hand, $k_a(B(t)) \leq k_a(t)$ holds due to the fact that $B(t) \leq t$.

\end{proof}

Next, we show that by employing the dynamic doubling batching scheme, BATS algorithm achieves optimal performance using only logarithmic rounds of communication (measured in terms of batches).

\banditBatches*
\begin{proof} [Proof of Theorem~\ref{thm:bandit_log_batch}]
Denote by $B_k$ the starting time of the $k$-th batch, and let $l_a(B_k)$ be the trigger integer for arm $a$ at time $B_k$, $K$ be the total number of rounds to interact with environment, namely, batches. Then for each arm $a \in \mathcal{A}$, $k_a(T) \leq T$, and
\begin{align*}
    k_a(T) = \sum_{k=1}^{K-1} k_a(B_{k+1}) - k_a(B_{k}) \leq \sum_{k=1}^{K-1} k_a(B_{k}) = \sum_{k=1}^{K-1} 2^{l_a(B_k) - 1} \leq \sum_{l=0}^{K-1} 2^{l},
\end{align*}
where the second and third step result from the dynamic batching scheme. Thus for each arm $a$, we have
\[
    K \leq \log(T+1).
\]
The proof is then completed by multiplying the above result by $N$ arms.
\end{proof}

\subsection{Regret Proofs in Multi-armed Bandit}
\label{sec:appendix_mab_regret}
With the convergence properties shown in Appendix~\ref{sec:appendix_sgld}, we proceed to prove the regret guarantee of Langevin TS with SGLD. 
The general idea of our regret proof is to upper bound the total number of times that the sub-optimal arms are pulled over time horizon $T$. We remark that the dependence of approximate samples  across batches complicates our analysis of TS compared to the existing analyses in bandit literature. 

We first decompose the expected regret according to the events of concentration in approximate samples $\theta_{a, t}$ and the events of estimation accuracy in expected rewards of sub-optimal arms.

For approximate samples $\theta$, define event
$
    E_{\theta,a}(B_{k}) = \lrbb{ \lrn{\theta_{a, k} - \theta_a^*} < C(n_k) },
$
which is guaranteed to happen with probability at least $(1 - \delta_2)$ by Lemma~\ref{lem:sgld_concentration} for some $\delta_2 \in [0, 1]$. Let $E_{\theta,a}(T) = \bigcap_{t=1}^{T}E_{\theta,a}(t)$, $E_{\theta,a}(K) = \bigcap_{k=1}^{K}E_{\theta,a}(B_k)$, where $K$ is the total number of batches. Without loss of generality, we
take $\lrn{\alpha_a}=1$ for all arms in $\mathbb{E}_{X \sim p_a(X|\theta_a)} [X] = \alpha_a^\intercal ~ \theta_a \leq \lrn{\theta_a}$ in the subsequent proofs

Let $\hat{\mu}_a(t)$ be the estimate of the expected reward for arm $a$ at time step $t$, and denote the filtration up to time $B(t)$ as $\mathcal{F}_{B(t)} := \lbrace a(\tau), r_{a(\tau), k_a(B(\tau))} \ | \ \tau \leq B(t) \rbrace$. For any sub-optimal arm $a \neq 1$, define event $E_{\mu,a}(t) = \{ \hat{\mu}_a(t) \geq \mu_1 - \epsilon \}$  with probability $p_{a, k_a(B(t))}(t) := \Prob(\hat{\mu}_a(t) \geq \mu_1 - \epsilon | \mathcal{F}_{B(t)})$ for some $\epsilon \in (0, 1)$, which signifies the estimation of arm $a$ is close to the true optimal expected reward.

\begin{lemma} [Regret Decomposition]
\label{lem:bandit_reg_parts}
Let $\mu_a$ be the true expected reward of arm $a$, $\mu^* = \max_{a \in \mathcal{A}} \mu_a$, $\Delta_a := \mu^* - \mu_a$. The expected regret of Langevin TS with SGLD satisfies:
\begin{align*}
    R_T \leq \sum_{a \in \mathcal{A}}  \Big( R_1 + R_2 + 2 \Big) \Delta_a, 
\end{align*}
where \resizebox{.9\hsize}{!}{$R_1 := \E{\sum_{t=1}^T \Ind(a(t) = a, E_{\mu,a}^c(t)) \ | \ E_{\theta,a}(K) \cap E_{\theta,1}(K)}, R_2 := \E{\sum_{t=1}^T \Ind(a(t) = a, E_{\mu,a}(t)) \ | \  E_{\theta,a}(K) \cap E_{\theta,1}(K)}$}.
\end{lemma}

\begin{proof}
[Proof of Lemma \ref{lem:bandit_reg_parts}.]
Recall that 
\begin{align*}
    R_T = \sum_{a \in \mathcal{A}} \Delta_a \cdot \mathbb{E} \lrb{k_a(T)}, \quad \Delta_a = \mu^* - \mu_a.
\end{align*}
For any sub-optimal arm $a \neq 1$, consider the event space $\F_\theta = \lbrace \lbrace E_{\theta,a}(T) \cap E_{\theta,1}(T) \rbrace, \lbrace E_{\theta,a}(T) \cap E_{\theta,1}(T) \rbrace^C \rbrace$, in which $E_{\theta,a}(T) \cap E_{\theta,1}(T)$ denotes the event that all approximate samples of arm $a$ and optimal arm $1$ are concentrated.

To bound the regret, we bound the largest number of times that each sub-optimal arm will be played:
\begin{align*}
    \E{k_a(T)} 
    &= \E{\sum_{t=1}^T \Ind(a(t) = a, E_{\mu,a}(t) \cup E_{\mu,a}^c(t), E_{\theta,a}(T) \cap E_{\theta,1}(T))} + \E{\sum_{t=1}^T \Ind(a(t) = a,  \lrp{E_{\theta,a}(T) \cap E_{\theta,1}(T) }^c)} \\
    &\leq \E{\sum_{t=1}^T \Ind(a(t) = a, E_{\mu,a}(t) \cup E_{\mu,a}^c(t)) \Big| E_{\theta,a}(T) \cap E_{\theta,1}(T)} 
    + \E{\sum_{t=1}^T \Ind(a(t) = a,  \lrp{E_{\theta,a}(T) \cap E_{\theta,1}(T) }^c)}.
\end{align*}
where the second inequality results from $P(E_{\theta,a}(T) \cap E_{\theta,1}(T)) \leq 1$.

For any arm $a \in \mathcal{A}$ in each batch, approximate samples are independently generated from the identical approximate distribution $\tilde{\rho}_a(\theta_a | R_a)$. Thus, approximate samples for arm $a$ are independent within the same batch, while being dependent across different batches, implying
\begin{align*}
    \begin{cases}
    \Prob(E_{\theta,a}(T))
    &= \prod_{t = 1}^T \Prob(E_{\theta,a}(t) | E_{\theta,a}(1), \dots, E_{\theta,a}(t-1)) = \prod_{k = 1}^{K-1} \Prob(E_{\theta,a}(B_{k+1}) | E_{\theta,a}(B_{k}))^{T_{k+1}} \\
    \Prob(E^c_{\theta,a}(T))
    &= \Prob(\bigcup_{t=1}^T E^c_{\theta,a}(t)) = \sum_{t = 1}^T \Prob(E^c_{\theta,a}(t)) = \sum_{k = 1}^K T_k \Prob(E^c_{\theta,a}(B_k))
    \end{cases}
    ,
\end{align*}
where $T_k := B_{k+1} - B_{k}$ is the number of time steps in the $k$-th batch, namely, the length of the batch. By Lemma~\ref{lem:sgld_concentration}, for each arm $a$ in batch $B_k$, $\Prob(E^c_{\theta,a}(B_k)) \leq \delta_2, (1 - \delta_2) \leq \Prob(E_{\theta,a}(B_k)) \leq 1$, which gives:
\begin{align*}
    \begin{cases}
        \Prob [ E_{\theta,a}(T) \cap E_{\theta,1}(T) ] \leq \Prob [ E_{\theta,a}(K) \cap E_{\theta,1}(K) ] \\
        \Prob [ E_{\theta,a}^c(T) \cup E_{\theta,1}^c(T) ] \leq \Prob [ E_{\theta,a}^c(T) ] + \Prob [ E_{\theta,1}^c(T) ] \leq 2 \delta_2 \sum_{k=1}^T T_k = 2 \delta_2 T \qquad\qquad\qquad\qquad\qquad
    \end{cases}
    .
\end{align*}
 Setting $\delta_2 = 1 / T^2$ gives,
\begin{align*}
\E{\sum_{t=1}^T \Ind(a(t) = a, \lrp{E_{\theta,a}(T) \cap E_{\theta,1}(T) }^c)} 
&= \E{ \sum_{t=1}^T \Ind(a(t) = a) \ | \ E_{\theta,a}(T)^c \cup E_{\theta,1}^c(T) } \Prob \bigg[{ E_{\theta,a}^c(T) \cup E_{\theta,1}^c(T) } \bigg] \\
& \leq 2 \delta_2 T \mathbb{E} \bigg[k_a(T) \ | \  E_{\theta,a}(T)^c \cup E_{\theta,1}(T)^c \bigg] \leq 2 \delta_2 T^2 \leq 2.
\end{align*}
Plugging in the results to the definition of regret yields,
\begin{align*}
    R(T) 
    &\leq \sum_{a \in \mathcal{A}} \Delta_a \cdot \Bigg( \E{\sum_{t=1}^T \Ind(a(t) = a, E_{\mu,a}(t) \cup E_{\mu,a}^c(t)) \ \Big| \ E_{\theta,a}(T) \cap E_{\theta,1}(T)} + 2 \Bigg) \\
    &\leq \sum_{a \in \mathcal{A}} \Delta_a \cdot \Bigg( \E{\sum_{t=1}^T \Ind(a(t) = a, E_{\mu,a}(t) \cup E_{\mu,a}^c(t)) \ \Big| \ E_{\theta,a}(K) \cap E_{\theta,1}(K)} + 2 \Bigg) \\
    &\resizebox{.95\hsize}{!}{$\leq \sum\limits_{a \in \mathcal{A}} \Delta_a \cdot \Bigg( \underbrace{\E{\sum_{t=1}^T \Ind(a(t) = a, E_{\mu,a}(t)) \ \Big| \ E_{\theta,a}(K) \cap E_{\theta,1}(K)}}_{R_1} + \underbrace{\E{\sum_{t=1}^T \Ind(a(t) = a,  E_{\mu,a}^c(t)) \ \Big| \ E_{\theta,a}(K) \cap E_{\theta,1}(K)}}_{R_2} + 2 \Bigg).$}
\end{align*}

\end{proof}

We then proceed to bound $R_1$ and $R_2$ respectively in Lemma~\ref{lem:bandit_reg_r1} and Lemma~\ref{lem:bandit_reg_r2}, the key to maintaining optimal regret is to maximize the probability of pulling the optimal arm by ensuring the event $E_{\mu,a}(t)$ takes place with low probability for all sub-optimal arms. 

\begin{lemma} [Bound term \textit{$R_1$}] 
\label{lem:bandit_reg_r1}
It can be shown that,
\[
    R_1 \leq \E{\sum_{t=1}^T \bigg( \frac{1}{p_{1,k_1(B(t))}(t)} - 1 \bigg) \ \bigg| \ E_{\theta,1}(K) }.
\]
    
\end{lemma}

\begin{proof} [Proof of lemma \ref{lem:bandit_reg_r1}.]
Note that arm $a$ is played at time $t$ if and only if $\hat{\mu}_{a'} \leq \hat{\mu}_{a}, \ \forall {a' \in \mathcal{A}}$. Thus for a sub-optimal arm $a$, the following event relationship holds: $\lbrace a(t) = a, E_{\mu,a}^c(t) \rbrace = \lbrace a(t) = a, E_{\mu,a}^c(t), \cap_{a^\prime \neq a} E_{\mu,a^\prime}^c(t) \rbrace \subseteq \lbrace \cap_{a^\prime \in \mathcal{A}} E_{\mu,a^\prime}^c(t) \rbrace $, and $\lbrace E_{\mu,1}(t), \cap_{a^\prime \neq 1} E_{\mu,a^\prime}^c(t) \rbrace = \lbrace a(t) = 1, E_{\mu,1}(t), \cap_{a^\prime \neq 1} E_{\mu,a^\prime}^c(t) \rbrace \subseteq \lbrace a(t) = 1 \rbrace $. We then have,
\begin{align*}
    \begin{cases}
    &\Prob \lrb{a(t) = a, E_{\mu,a}^c(t) \ | \ \mathcal{F}_{B(t)}}
    \leq \Prob \big[\bigcap_{a^\prime \in \mathcal{A}} E_{\mu,a^\prime}^c(t) \ | \ \mathcal{F}_{B(t)} \big]
    = \Prob \big[\bigcap_{a^\prime \neq 1} E_{\mu,a^\prime}^c(t) \ | \ \mathcal{F}_{B(t)} \big] \big(1 - \Prob \big[ E_{\mu,1}(t) \ | \ \mathcal{F}_{B(t)} \big] \big) \\
    &\Prob \lrb{a(t) = 1 \ | \ \mathcal{F}_{B(t)}}
    \geq \Prob \big[ E_{\mu,1}(t) \ | \ \mathcal{F}_{B(t)} \big] \Prob \big[\bigcap_{a^\prime \neq 1} E_{\mu,a^\prime}^c(t) \ | \ \mathcal{F}_{B(t)} \big] 
    \end{cases} 
    \end{align*}
    
Recall that $p_{1,k_1(B(t))}(t)  := \Prob [ E_{\mu,1}(t) \ | \ \mathcal{F}_{B(t)} ]$. Combining the above two equations shows that the probability of pulling a sub-optimal arm $a$ is bounded by the probability of pulling the optimal arm with an exponentially decaying coefficient:
\begin{equation}
    \Prob \lrb{a(t) = a, E_{\mu,a}^c(t) \ | \ \mathcal{F}_{B(t)}} \leq \bigg( \frac{1}{p_{1,k_1(B(t))}(t)} - 1 \bigg) \Prob \lrb{a(t) = 1 \ | \ \mathcal{F}_{B(t)}}.
\end{equation}

Therefore, $R_1$ is upper bounded accordingly:
    \begin{align*}
        R_1 
        &=\E{\sum_{t=1}^T \E{\Ind(a(t) = a, E_{\mu,a}^c(t) \ | \ \mathcal{F}_{B(t)}} \ \bigg| \ E_{\theta,a}(K) \bigcap E_{\theta,1}(K) } \\
        &= \E{\sum_{t=1}^T \Prob \lrb{a(t) = a, E_{\mu,a}^c(t) \ | \ \mathcal{F}_{B(t)}} \ \bigg| \ E_{\theta,a}(K) \bigcap E_{\theta,1}(K) } \\
        &\leq \E{\sum_{t=1}^T \bigg( \frac{1}{p_{1,k_1(B(t))}(t)} - 1 \bigg) \Prob \lrb{a(t) = 1 \ | \ \mathcal{F}_{B(t)}} \ \bigg| \ E_{\theta,a}(K) \bigcap E_{\theta,1}(K) } \\
        &= \E{\sum_{t=1}^T \bigg( \frac{1}{p_{1,k_1(B(t))}(t)} - 1 \bigg) \Ind \big[ a(t) = 1 \big] \ \bigg| E_{\theta, 1}(K) } \\
        &\leq \E{\sum_{t=1}^T \bigg( \frac{1}{p_{1,k_1(B(t))}(t)} - 1 \bigg) \ \bigg| \ E_{\theta,1}(K) }.
    \end{align*}
    
\end{proof}

\begin{lemma} [Bound term \textit{$R_2$}]
\label{lem:bandit_reg_r2}
It can be shown that,
\[
    R_2 \leq 1 + \E{\sum_{t=1}^{T}\Ind \lrp{p_{a, k_a(B(t))}(t) > \frac{1}{T}}  \ \bigg| \ E_{\theta,a}(K) }.
\]
    
\end{lemma}

\begin{proof}[Proof of Lemma \ref{lem:bandit_reg_r2}.]
The proof closely follows \cite{agrawal2012analysis}. Let $\mathcal{T} := \lbrace t \ | \ p_{a, k_a(B(t))}(t) > \frac{1}{T} \rbrace$. $R_2$ term can be rewritten as: 
\begin{align*}
        R_2
        &= \E{\sum_{t \in \mathcal{T}} \Ind(a(t) = a, E_{\mu,a}(t)) \ \bigg| \  E_{\theta,a}(K) \bigcap E_{\theta,1}(K)} + \E{\sum_{t \notin \mathcal{T}} \Ind(a(t) = a, E_{\mu,a}(t)) \ \bigg| \  E_{\theta,a}(K) \bigcap E_{\theta,1}(K)} \\
        &\leq \underbrace{\E{\sum_{t \in \mathcal{T}} \Ind(a(t) = a) \ \bigg| \  E_{\theta,a}(K) \bigcap E_{\theta,1}(K)}}_{I} + \underbrace{\E{\sum_{t \notin \mathcal{T}} \Ind(E_{\mu,a}(t)) \ \bigg| \  E_{\theta,a}(K) \bigcap E_{\theta,1}(K)}}_{II}. 
\end{align*}
    
It follows that the first term satisfies,
\begin{align*}
    I = \E{\sum_{t = 1}^T \Ind(a(t) = a, p_{a, k_a(B(t))}(t) > \frac{1}{T} ) \ \bigg| \  E_{\theta,a}(K)} \leq \E{\sum_{t = 1}^T \Ind(p_{a, k_a(B(t))}(t) > \frac{1}{T} ) \ \bigg| \  E_{\theta,a}(K)},
    \end{align*}
    
and the second term satisfies,
\begin{align*}
    II = \E{\sum_{t \notin \mathcal{T}} \mathbb{E} \Big[\Ind(E_{\mu,a}(t)) \ \big | \ \mathcal{F}_{B(t)} \Big] \ \bigg| \  E_{\theta,a}(K) \bigcap E_{\theta,1}(K)} =  \E{\sum_{t \notin \mathcal{T}} p_{a, k_{a}(B(t))}(t) \ \bigg| \  E_{\theta,a}(T) \bigcap E_{\theta,1}(T)} \leq 1,
    \end{align*}
where the last inequality holds as $p_{a, k_a(B(t))}(t) \leq 1 / T$ for $t \notin \mathcal{T}$.
\end{proof}

\begin{lemma}
\label{lem:bandit_anti_concentration}
Assume that the prior and reward distributions satisfy Assumptions \ref{ass:1}-\ref{ass:4}. Then at each time step $t \leq T$, if there are $k_1(B(t))$ observed rewards for arm $1$, then Algorithm \ref{alg:TS_SGLD} ensures:
\[
    \E{\frac{1}{p_{1,k_1(B(t))}(t)}} \leq 36 \sqrt{Q_1},
\]
where $Q_1 = \max_{\theta \in \R^d} \frac{p_{1}(\theta)}{p_{1}(\theta_1^*)}$ measures the quality of the prior distribution, $Q_1 \geq 1$.
\end{lemma}

\begin{proof} [Proof of Lemma \ref{lem:bandit_anti_concentration}]
For completeness, we provide the proof of this lemma, which closely follows the proof of Lemma 18 in \cite{mazumdar2020thompson}.

For each arm $a$, upon running SGLD with batched data in batch $k$, by Cauchy-Schwartz inequality, we have,
\[
    \mathbb{P} \left( \alpha_a^{\rT}(\theta_a^k - \theta_{a, N\eta}) \geq \alpha_1^{\rT}(\theta_a^* - \theta_{a, N\eta}) -\epsilon \right) \geq \mathbb{P} \left( Z \geq \lrn{\theta_a^* - \theta_{a, N\eta}} \right),
\]
where $Z \sim \mathcal{N}(0, \frac{1}{nL\gamma}I)$. Let $\sigma^2 = \frac{1}{nL\gamma}I$, by anti-concentration of Gaussian random variables, for the optimal arm $1$,
\begin{align*}
    p_{1,k_1(B(t))}(t) \geq \sqrt{\frac{1}{2\pi}} \begin{cases}
        \frac{\sigma t}{t^2 + \sigma^2}e ^{-\frac{t^2}{2\sigma^2}}, &~~t > \sigma; \\
        0.34, &~~\text{otherwise}.
    \end{cases}
\end{align*}
Taking expectations of both sides and by Cauchy-Schwartz inequality,
\begin{align*}
    \E{\frac{1}{p_{1,k_1(B(t))}(t)}} 
    &\leq 3\sqrt{2 \pi}+\sqrt{2\pi nL\gamma}\sqrt{\mathbb{E} \left[\|\theta_1^* - \theta_{1,Nh}\|^2\right]}\sqrt{\mathbb{E} \left[e^{nL\gamma \|\theta_1^* - \theta_{1,Nh}\|^2}\right]}+\sqrt{2\pi}\mathbb{E}\left[e^{\frac{nL\gamma}{2} \|\theta_1^* - \theta_{1,Nh}\|^2} \right]. 
\end{align*}
By the convergence guarantee of SGLD in  Theorem~\ref{thm:sgld_convergence_app},
\[
    \mathbb{E} \left[\|\theta_1^* - \theta_{1,Nh}\|^2 \right] \leq \frac{18}{mn}\left(d+\log Q+32+ \frac{8dL^2}{\nu m} \right). 
\]
Note that $\|\theta_1^* - \theta_{1,Nh}\|^2$ is a sub-Gaussian random variable, when $\gamma \leq \frac{m}{32L\sigma}$, 
\[
    \mathbb{E}[e^{nL\gamma\|\theta_1^* - \theta_{1,Nh}\|^2}] \leq 3 / 2\left(e^{\frac{4nL\gamma D}{m}} +2.5\right).
\]
Combining the above results together completes the proof.
\end{proof}

With Lemma \ref{lem:bandit_anti_concentration}, we now proceed to prove the terms in $R_1$ and $R_2$ that lead us to the final regret bound.
\begin{lemma}
\label{lem:bandit_reg_sums}
Assume that Assumptions 1-4 are satisfied. Let $ \sigma = 16 + \frac{4dL^2}{\nu m}$, $\gamma = \frac{m}{32L\sigma}$. running Algorithm \ref{alg:TS_Bandit} with samples generated from approximate posteriors using Algorithm \ref{alg:TS_SGLD}, we have, 
\begin{align}
    \E{\sum_{t=1}^T \bigg( \frac{1}{p_{1,k_1(B(t))}(t)} - 1 \bigg) \ \bigg| \ E_{\theta,1}(K) }\leq   \frac{20736e}{m\Delta_a^2} \sqrt{Q_1} \big( d + \log Q_1 + 4\sigma \log T + 12d \sigma \log 2 \big) + 1.
\end{align}

\begin{align}
    \E{\sum_{t=1}^{T}\Ind \lrp{p_{a, k_a(B(t))}(t) > \frac{1}{T}}  \ \bigg| \ E_{\theta,a}(K) } \leq \frac{576e}{m \Delta_a^2} \big( d + \log Q_a + 10d\sigma \log (T) \big). 
\end{align}

\end{lemma}

\begin{proof} [Proof of lemma \ref{lem:bandit_reg_sums}.]

For ease of notation, let the number of observed rewards in batch $k$ for arm $a \in \mathcal{A}$ be $n_k$. By definition,
\begin{align*}
    \quad p_{1, n_k} 
    = \mathbb{P} \big( \hat{\mu}_1(t) \geq \mu_1 - \epsilon \ \big | \mathcal{F}_{B(t)} \big) \geq 1 - \mathbb{P} \big( \lrn{\theta_1 - \theta_1^*} > \epsilon \ \big | \mathcal{F}_{B(t)} \big) 
\end{align*}
With concentration property of approximate samples in Lemma ~\ref{lem:sgld_concentration}, it suggests the increasing number of observed rewards for the optimal arm leads to the increasing probability of being optimal. Thus by Lemma ~\ref{lem:bandit_pull_relation}, at any time step $t \leq T $, 
\[
    p_{1, k_1(B(t))} (t) \geq p_{1, \frac{k_1(t)}{2}} (t).
\]
Concentration is achieved only when sufficient number of rewards is observed, we thus require:
\begin{equation}
    \label{SGLD_con}
    \mathbb{P}_{\theta_1 \sim \tilde{\rho}_{1, n_k}[\gamma]} \lrp{ \lrn{\theta_1 - \theta_1^*} \geq \epsilon } \leq \exp \lrp{ -\frac{1}{6d\sigma}\lrp{ \frac{m n_k\epsilon^2}{36e} - \Bar{D}_1 }},
\end{equation}
where $\Bar{D}_1 = d + \log Q_1 + 4\sigma \log T, \sigma = 16 + \frac{4dL^2}{\nu m}$. Choose $\epsilon = (\mu_1 - \mu_a) / 2 = \Delta_a / 2$, and consider the time step $t$ when arm $1$ satisfies: 
\[
    k_1(t) = 2^{\lceil \log_2 2l \rceil}, \qquad where \quad l =  \frac{144e}{m \Delta_a^2}\big( \Bar{D}_1 + 6d \sigma \log 2 \big).
\]
As $k_1(t) \geq 2l$, the number of observed rewards is guaranteed to be at least $ \frac{36e}{m\epsilon^2}\Bar{D}$, and $\mathbb{P}_{\theta_1 \sim \tilde{\rho}_{1,n_k}[\gamma]} ( \lrn{\theta_1 - \theta_1^*} $ $\geq \epsilon )\leq 1/2$.
Thus, the individual term in $R_1$ follows:
\begin{align*}
    &\quad \E{\sum_{t=1}^T \bigg( \frac{1}{p_{1,k_1(B(t))}(t)} - 1 \bigg) \ \bigg| \ E_{\theta,1}(K) } \\
    &\leq \E{\sum_{t=1}^T \bigg( \frac{1}{p_{1,\frac{k_1(t)}{2}}(t)} - 1 \bigg) \ \bigg| \ E_{\theta,1}(K) } \\
    &\leq \E{\sum_{k_1(t)=0}^{T-1} \bigg( \frac{1}{p_{1,\frac{k_1(t)}{2}}(t)} - 1 \bigg) \ \bigg| \ E_{\theta,1}(K) } \\
    &\leq \E{\sum_{k_1(t)=0}^{2^{\lceil \log_2 2l \rceil}} \bigg( \frac{1}{p_{1,\frac{k_1(t)}{2}}(t)} - 1 \bigg) \ \bigg| \ E_{\theta,1}(K) } + \E{\sum_{k_1(t)=2^{\lceil \log_2 2l \rceil}+1}^{T-1} \bigg( \frac{1}{p_{1,\frac{k_1(t)}{2}}(t)} - 1 \bigg) \ \bigg| \ E_{\theta,1}(K) }. \numberthis
    \label{eqn:bandit_r1_term}
\end{align*}

In early stage when concentration has not been achieved, using results from Lemma~\ref{lem:bandit_anti_concentration},
\begin{equation}
    \label{eqn:bandit_before_con}
    \E{\sum_{k_1(t)=0}^{2^{\lceil \log_2 2l \rceil}} \bigg( \frac{1}{p_{1,\frac{k_1(t)}{2}}(t)} - 1 \bigg) \ \bigg| \ E_{\theta,1}(K) } \leq 2^{\lceil \log_2 2l \rceil} 36 \sqrt{B_1} \leq 2 \cdot 2l \cdot 36 \sqrt{B_1}.
\end{equation}

When sufficient rewards for the optimal arm has been accumulated, 
\begin{align*}
    &\quad\E{\sum_{k_1(t)=2^{\lceil \log_2 2l \rceil}+1}^{T-1} \bigg( \frac{1}{p_{1,\frac{k_1(t)}{2}}(t)} - 1 \bigg) \ \bigg| \ E_{\theta,1}(K) } \\
    &\leq \E{\sum_{k_1(t)=0}^{T-1} \bigg( \frac{1}{p_{1,\frac{k_1(t)}{2}}(t)} - 1 \bigg) \ \bigg| \ E_{\theta,1}(K) } \\
    &\leq \sum_{k_1(t)=0}^{T-1}  \frac{1}{\exp \lrp{ -\frac{1}{6d\sigma_1}\lrp{ \frac{m_1 \epsilon^2}{36e} \cdot \frac{k_1(t)}{2} }}} - 1 \\
    &\leq \int_{z=0}^{\infty} \lrp{ \frac{1}{\exp \lrp{ -\frac{m \epsilon^2}{432ed\sigma_1} z }} - 1 } dz \\
    &\leq 2 \cdot \frac{144e }{m \Delta_a^2} \cdot 6d\sigma \log 2 + 1. \numberthis \label{eqn:bandit_after_con}
\end{align*}

Substituting equation (\ref{eqn:bandit_before_con}) and (\ref{eqn:bandit_after_con}) back to (\ref{eqn:bandit_r1_term}) yields,
\begin{align*}
    \E{\sum_{t=1}^T \bigg( \frac{1}{p_{1,k_1(B(t))}(t)} - 1 \bigg) \ \bigg| \ E_{\theta,1}(K) } 
    &\leq 4l \cdot 36 \sqrt{Q_1} + 2 \cdot \frac{144e }{m \Delta_a^2} \cdot 6d\sigma \log 2 + 1 \\
    &\leq 36 \sqrt{Q_1} \frac{576e}{m\Delta_a^2}\big( \Bar{D}_1 + 12d \sigma\log 2 \big) + 1.
\end{align*}

Similarly, for $R_2$ term with event $E_{\mu,a}(t) = 
\lbrace \hat{\mu}_a(t) \geq \mu_1 - \epsilon \rbrace$, let $\epsilon = (\mu_1 - \mu_a) / 2 = \Delta_a / 2$,
\begin{align*}
    p_{a, k_a(B(t))}(t) 
    &= \Prob(\hat{\mu}_a(t) - \mu_a \geq \mu_1 - \mu_a - \epsilon | \mathcal{F}_{B(t)}) \\
    &= \Prob(\hat{\mu}_a(t) - \mu_a \geq \frac{\Delta_a}{2} | \mathcal{F}_{B(t)}) \\
    &\leq  \Prob(\hat{\mu}_a(t) - \mu_a \geq  \frac{\Delta_a}{2} | \mathcal{F}_{\frac{k_a(t)}{2}}) \\
    &= p_{a, \frac{k_a(t)}{2}}(t),
\end{align*}
which gives
\begin{align*}
    &\quad \E{\sum_{t=1}^{T}\Ind \lrp{p_{a, k_a(B(t))}(t) > \frac{1}{T}}  \ \bigg| \ E_{\theta,a}(K) } \\
    &\leq \E{\sum_{t=1}^{T}\Ind \lrp{ \Prob(\hat{\mu}_a(t) - \mu_a \geq \frac{\Delta_a}{2} | \mathcal{F}_{\frac{k_a(t)}{2}}) > \frac{1}{T} }  \ \bigg| \ E_{\theta,a}(K) } \\
    &\leq \E{\sum_{t=1}^{T}\Ind \lrp{ \Prob( |\hat{\mu}_a(t) - \mu_a | \geq \frac{\Delta_a}{2} | \mathcal{F}_{\frac{k_a(t)}{2}}) > \frac{1}{T} }  \ \bigg| \ E_{\theta,a}(K) } \\
    &\leq \E{\sum_{t=1}^{T}\Ind \lrp{ \mathbb{P}_{\theta_a \sim \tilde{\rho}_{a,\frac{k_a(t)}{2}}}  \big( \lrn{\theta_a - \theta_a^*} \geq \frac{\Delta_a}{2} \big) > \frac{1}{T} }  \ \bigg| \ E_{\theta,a}(K) }.
\end{align*}

With the same form of posterior as in equation \ref{SGLD_con}, $\mathbb{P}_{\theta_a \sim \tilde{\rho}_{a,\frac{k_a(t)}{2}}}  \big( \lrn{\theta_a - \theta_a^*} \geq \frac{\Delta_a}{2} \big) \leq \frac{1}{T}$ for arm $a \neq 1$ holds, when 
\[
    k_a(t) > 2\cdot 2 \cdot \frac{144e}{m \Delta_a^2} \big( \Bar{D}_a + 6d \sigma \log (T) \big).
\]
Here, the number of observed rewards is guaranteed to be at least $2^{\lceil \log_2 l \rceil}$, where $l = \frac{144e}{m\Delta_a^2} \big( \Bar{D}_a + 6d \sigma \log (T) \big)$.
Therefore, using the fact that $d > 1$, we have,
\[
    \E{\sum_{t=1}^{T}\Ind \lrp{p_{a, k_a(B(t))}(t) > \frac{1}{T}}  \ \bigg| \ E_{\theta,a}(K) } \leq \frac{576e}{m_a \Delta_a^2} \big( \Bar{D}_a + 6d\sigma_a \log (T) \big).
\]

\end{proof}

We are ready to prove the final regret bound by combining results from the above Lemmas.
\banditRegret*

\begin{proof}
[Proof of Theorem \ref{thm:reg_batch_SGLD}.]
The proof is a direct result by combining Lemma~\ref{lem:bandit_reg_parts}, \ref{lem:bandit_reg_r1}, \ref{lem:bandit_reg_r2},\ref{lem:bandit_reg_sums},
which gives, 
\begin{align*}
    R_T 
    &\leq \sum_{a \in \mathcal{A}} \Delta_a \cdot \Bigg( R_1 + R_2 + 2 \Bigg) \\
    &\leq \Bigg( \sum_{a  \in \mathcal{A}} 4 \cdot 36 \sqrt{Q_1} \frac{144e}{m\Delta_a}\big( d + \log Q_1 + 4\lrp{16 + \frac{4dL^2}{m \nu}} \lrp{\log T + 3d\log 2} \big) + \Delta_a \\
    &+ \Delta_a + 4 \frac{144e}{m\Delta_a} \big( d + \log Q_a + 10d \lrp{16 + \frac{4dL^2}{m \nu}} \log (T) \big) + 2\Delta_a \Bigg)\\
    &\leq \sum_{a > 1} \frac{C \sqrt{Q_1} }{m\Delta_a} \lrp{d + \log Q_1 + d\kappa^2 \log T + d^2 \kappa^2 } + \frac{C}{m \Delta_a} \lrp{d + \log Q_a +  d^2 \kappa^2 \log T } + 4\Delta_a.
\end{align*}

\end{proof}

\section{Proofs of Langevin Posterior Sampling for Reinforcement Learning} \label{app:mdp}

\allowdisplaybreaks

In this section, we will present the regret proofs for Langevin Posterior Sampling algorithms in RL frameworks under different types of parameterization, and conclude with a real-world example where the General Parameterization from Section \ref{sec:general_mdp} is applicable. 

\subsection{Communication cost of Static Doubling Batching Scheme}

\counterwithin{lemma}{section}

We first show that under the static doubling batching scheme in RL setting, LPSRL algorithm achieves optimal performance using only logarithmic rounds of communication (measured in terms of batches, or equivalently policy switches).

\begin{theorem}
\label{thm:mdp_communication_cost}
   Let $T_k$ be the number of time steps between the $(k-1)$-th policy switch and the $k$-th policy switch, and $K_T$ be the total number of policy switches for time horizon $T$. LPSRL ensures that 
   \[
        K_T \leq \log T + 1. 
   \]
\end{theorem}

\begin{proof} [Proof of Theorem \ref{thm:mdp_communication_cost}.]
    By design of Algorithm~\ref{alg:TS_alg_MDP_double},at the $k$-th policy switch, $T_k = 2^{k-1}$. Since the total number of time steps is determined by time horizon $T$, we can easily obtain $K_T = \lceil \log T \rceil$.
\end{proof}

\subsection{Regret Proofs in Average-reward MDPs} \label{app:mdp_proofs}

In this section, we proceed to prove the theorems in Section \ref{sec:mdp}. To focus on the problem of model estimation, our results are developed under the optimality of policies\footnote{If only suboptimal policies are available in our setting, it can be shown that small approximation errors in policies only contribute additive non-leading terms to regret. See details in \cite{ouyang2017learning}.}. 

While analyses of Bayes regret in existing works of PSRL crucially depend on the true transition dynamics $\theta^*$ being identically distributed as those of sampled MDP \cite{russo2014learning}, we show that in Langevin PSRL, sampling from the approximate posterior instead of the true posterior will introduce a bias that can be upper bounded using Wasserstein-$1$ distance.

\LangevinPS*

\begin{proof} [Proof of Lemma~\ref{lem:MDP_PS_main_change}]
Notice that both $\tilde{\rho}_{t_k}$ and ${\rho}_{t_k}$ are measurable with respect to $\sigma(\mathcal{H}_{t_k})$. Therefore, condition on history $\mathcal{H}_{t_k}$, the only randomness under the expectation comes from the sampling procedure for approximate posterior, which gives,
\begin{align*}
\expectation[f(\theta^k) | \mathcal{H}_{t_k}] &= \int_{\reals^d} f (\theta) \tilde{\rho}_{t_k} (d\theta)\\
&= \int_{\reals^d} f (\theta) (\tilde{\rho}_{t_k} - \rho_{t_k} + \rho_{t_k} - \delta(\theta^*) +  \delta(\theta^*)) (d\theta)\\
&\leq \expectation[f(\theta^{k,*}) | \mathcal{H}_{t_k}] - \expectation[f(\theta^{*}) | \mathcal{H}_{t_k}] + \expectation[f(\theta^{*}) | \mathcal{H}_{t_k}] + W_1(\tilde{\rho}_{t_k}, \rho_{t_k}) \\
&= \expectation[f(\theta^{*}) | \mathcal{H}_{t_k}] + W_1(\tilde{\rho}_{t_k}, \rho_{t_k}) \numberthis \label{eqn:mdp_key_prop_1}.
\end{align*}

The third inequality follows from the fact that given $\mathcal{H}_{t_k}$, $\rho_{t_k}$ is the posterior of $\theta^{k,*}$  and the definition of dual representation for $W_1$ with respect to the $1$-Lipschitz function $f$. The last equality follows from the standard posterior sampling lemma in the Bayesian setting \citep{osband2013more, osband2014model}, which suggests that at time $t_k$, given the sigma-algebra $\sigma(\mathcal{H}_{t_k})$, $\theta^{k,*}$ and $\theta^{*}$ are identically distributed:
\[
    \expectation[f(\theta^{k,*}) | \mathcal{H}_{t_k}] = \expectation[f(\theta^{*}) | \mathcal{H}_{t_k}].
\]
Following the same argument, condition on $\mathcal{H}_{t_k}$, we also have,
\begin{equation}
\label{eqn:mdp_key_prop_2}
    \expectation[f(\theta^{*}) | \mathcal{H}_{t_k}] = \expectation[f(\theta^{k,*}) | \mathcal{H}_{t_k}]
    = \int_{\reals^d} f (\theta) (\rho_{t_k} + \tilde{\rho}_{t_k} - \tilde{\rho}_{t_k}) (d\theta)
    \leq \expectation[f(\theta^k) | \mathcal{H}_{t_k}] + W_1(\tilde{\rho}_{t_k}, \rho_{t_k}).
\end{equation}
Combining Equation (\ref{eqn:mdp_key_prop_1}) and (\ref{eqn:mdp_key_prop_2}) yields Equation (\ref{eqn:mdp_key_prop}). Applying the tower rule concludes the proof.
\end{proof}

\begin{corollary} [Tabular Langevin Posterior Sampling]
\label{cor:MDP_tabular_PS_main_change}
In tabular settings with finite states and actions, 
by running an approximate sampling method for each $(s,a) \in \mathcal{S} \times \mathcal{A}$ at time $t_k$, it holds that for each policy switch $k \in [K_T]$,
\[
    \Big|  \expectation [f(\theta^*) | \mathcal{H}_{t_k}] - \expectation [f(\theta^k) | \mathcal{H}_{t_k}] \Big| \leq \sum_{(s,a) \in \mathcal{S} \times \mathcal{A}} W_1(\tilde{\rho}_{t_k}(s,a), \rho_{t_k}(s,a)),
\]
where 
$\tilde{\rho}_{t_k}(s,a)$ are the corresponding true posterior and approximate posterior for $(s, a)$ at time $t_k$.
\end{corollary}

\begin{proof} [Proof of Corollary~\ref{cor:MDP_tabular_PS_main_change}]
    Since we run the approximate sampling algorithm for each state-action pair at the beginning of each policy switch $k$, the total approximation error is equal to the sum of approximation error for each $(s, a)$.
\end{proof}

We first provide a general regret decomposition in Lemma \ref{lem:mdp_regret_decomp}, which holds for any undiscounted weakly-communicating MDPs with infinite horizon, where approximate sampling is adopted and the transition is Lipschitiz. 

\begin{lemma} [Regret decomposition.]
\label{lem:mdp_regret_decomp}
For a weakly-communicating MDP with infinite time-horizon $T$, the Bayesian regret of Algorithm \ref{alg:TS_alg_MDP_double} instantiated with any approximate sampling method can be decomposed as follows: 
\begin{equation}
\label{eqn:mdp_regret_decom}
    R_B(T) \leq \mathbb{E} \Big[ \sum_{k=1}^{K_T} T_k W_1(\tilde{\rho}_{t_k}, \rho_{t_k}) \Big] + H (\log T + 1) + H L_p \mathbb{E} \Big[ \sum_{k=1}^{K_T} \sum_{t = t_k}^ {t_{k+1}-1}  \lrn{\theta^* - \theta^k} \Big], 
\end{equation}
where $L_p$ is a Lipschitz constant, and $H$ is the upper bound of span of MDP.
\end{lemma}

\begin{proof} [Proof of Lemma~\ref{lem:mdp_regret_decomp}]
We adopt the greedy policy with respect to the sampled model, which gives $a_t = \argmax_{a \in A}$ $r(s_t, a)$ at each time step $t$. By Bellman Optimality equation in Lemma \ref{lem:bellman_mdp},
\begin{equation}
\label{eqn:bellman}
    J^{\pi_k}(\theta^k) + h^{\pi_k}(s, \theta^k) = \mathcal{R}(s_t, a_t) + \int_{s^\prime \in \mathcal{S}}p(s^\prime | s_t, a_t; \theta^k) h^{\pi_k}(s^\prime, \theta^k) d s^\prime, ~~~~\forall t \in [t_k, t_{k+1} - 1].
\end{equation}
We then follow the standard analyses in RL literature \cite{osband2013more, osband2014model} to decompose the regret into sum of Bellman errors. Plug in Equation~(\ref{eqn:bellman}) into the definition of Bayesian regret, we have,
\begin{align*}
    R_B(T) 
    &= \mathbb{E} \lrb{\sum_{t=1}^T J^{\pi^*}(\theta^*) - \mathcal{R}(s_t, a_t)} \\
    &= \mathbb{E} \lrb{\sum_{k=1}^{K_T} \sum_{t = t_k}^ {t_{k+1}-1} J^{\pi^*}(\theta^*) - \mathcal{R}(s_t, \pi_k(s_t))} \\
    &= \resizebox{0.83\hsize}{!}{$\underbrace{\mathbb{E} \Big[ \sum_{k=1}^{K_T} \sum_{t = t_k}^ {t_{k+1}-1} J^{*}(\theta^*) - J^{\pi_k}(\theta^k) \Big]}_{\text{(i)}} + \underbrace{\mathbb{E} \lrb{ \sum_{k=1}^{K_T} \sum_{t = t_k}^ {t_{k+1}-1} \lrp{ \int_{s^\prime \in \mathcal{S}} p(s^\prime | s_t, \pi_k(s_t); \theta^k)h^{\pi_k}(s^\prime, \theta^k) d s^\prime - h^{\pi_k}(s_t, \theta^k)}} }_{\text{(ii)}}  \numberthis \label{eqn:mdp_regret_decom_1}$ }
    \end{align*}

\textbf{Term (i)}. By the property of approximate posterior sampling in Lemma \ref{lem:MDP_PS_main_change} and the non-negativity of Wasserstein distance, 
\begin{align*}
  \mathrm{(i)} \leq | \mathrm{(i)} | 
  &\leq \mathbb{E} \Big[ \sum_{k=1}^{K_T} \sum_{t = t_k}^{t_{k+1}-1} \Big| J^{*}(\theta^*) - J^{\pi_k}(\theta^k) \Big| \Big] \leq \mathbb{E} \Big[ \sum_{k=1}^{K_T} \sum_{t = t_k}^ {t_{k+1}-1} W_1(\rho_{t_k}, \tilde{\rho}_{t_k}) \Big]  = \mathbb{E} \Big[ \sum_{k=1}^{K_T} T_k W_1(\tilde{\rho}_{t_k}, \rho_{t_k}) \Big]. \numberthis \label{eqn:mdp_regret_decom_2}
\end{align*}
We remark that this term differs from the exact PSRL where no approximate sampling method is used. To ensure the final regret is properly bounded, approximate sampling method being used must provide sufficient statistical guarantee of $\tilde{\rho}_{t_k}$ and $\rho_{t_k}$ in terms of Wasserstein-$1$ distance.

\textbf{Term (ii)}. We further decompose term (ii) into the model estimation errors.
\begin{align*}
    \mathrm{(ii)}
    &= \mathbb{E} \Big[ \sum_{k=1}^{K_T} \sum_{t = t_k}^ {t_{k+1}-1} \Big(\int_{s^\prime \in \mathcal{S}} p(s^\prime | s_t, \pi_k(s_t); \theta^k)h^{\pi_k}(s^\prime, \theta^k) d s^\prime - h^{\pi_k}(s_t, \theta^k) + h^{\pi_k}(s_{t+1}, \theta^k) - h^{\pi_k}(s_{t+1}, \theta^k) \Big) \Big] \\
    &= \underbrace{ \mathbb{E} \Big[ \sum_{k=1}^{K_T} \sum_{t = t_k}^ {t_{k+1}-1} \Big( h^{\pi_k}(s_{t+1}, \theta^k) - h^{\pi_k}(s_{t}, \theta^k) \Big) \Big] }_{\Delta_h} \\
    &+ \underbrace{ \mathbb{E} \Big[ \sum_{k=1}^{K_T} \sum_{t = t_k}^ {t_{k+1}-1} \int_{s^\prime \in \mathcal{S}} \Big(p(s^\prime | s_t, \pi_k(s_t), \theta^k) - p(s^\prime | s_t, \pi_k(s_t), \theta^*) \Big) h^{\pi_k}(s^\prime, \theta^k) d s^\prime \Big] }_{\Delta_{err}}. 
\end{align*}

To bound $\Delta_h$, note that for each $k \in [1, K_T]$, $sp(h(\theta^k)) \leq H$, and by Theorem \ref{thm:mdp_communication_cost}, 
\begin{align*}
   \Delta_h 
   &=  \mathbb{E} \Big[ \sum_{k=1}^{K_T} \sum_{t = t_k}^ {t_{k+1}-1} \Big( h^{\pi_k}(s_{t+1}, \theta^k) - h^{\pi_k}(s_{t}, \theta^k) \Big) \Big] \\
    &= \mathbb{E} \Big[ \sum_{k=1}^{K_T} \Big( h^{\pi_k}(s_{t_{k+1}}, \theta_k) - h^{\pi_k}(s_{t_k}, \theta_k) \Big) \Big] \\
    &\leq \E{sp(h(\theta^k)) K_T} \\
    &\leq H(\log T + 1) \numberthis \label{eqn:mdp_regret_decom_3}.
\end{align*}

Thus, combining Equation (\ref{eqn:mdp_regret_decom_1}),(\ref{eqn:mdp_regret_decom_2}), (\ref{eqn:mdp_regret_decom_3}), and by Lemma~
\ref{lem:mdp_regret_term3_bound}, we conclude the proof.

\end{proof}



\begin{lemma} [Bound estimation error] 
\label{lem:mdp_regret_term3_bound}
Let $\Delta_{err} = \mathbb{E} \Big[ \sum_{k=1}^{K_T} \sum_{t = t_k}^ {t_{k+1}-1} \int_{s^\prime \in \mathcal{S}} \Big(p(s^\prime | s_t, \pi_k(s_t), \theta^k) - p(s^\prime | s_t, \pi_k(s_t),$ $\theta^*) \Big) h^{\pi_k}(s^\prime, \theta^k) d s^\prime \Big]$. Suppose Assumption~\ref{ass:mdp_lip_transition} holds, then

\begin{equation}
\label{eqn:mdp_delta_err_term_1}
    \Delta_{err} \leq H L_p \mathbb{E} \Big[ \sum_{k=1}^{K_T} \sum_{t = t_k}^ {t_{k+1}-1}  \lrn{\theta^* - \theta^k} \Big].
\end{equation}
\end{lemma}
\begin{proof} [Proof of Lemma \ref{lem:mdp_regret_term3_bound}]
Recall that $\pi_k$ is the optimal policy under $\theta^k$, thus $h(\cdot, \theta^k) = h^{\pi_k}(\cdot, \theta^k)$, and span is properly bounded in weakly-communicating MDPs: $sp(h(\theta)) \leq H$ for any $\theta \in \R^d$. Then by Assumption \ref{ass:mdp_lip_transition} and Cauchy-Schwartz inequality,
\begin{align*}
    &\qquad \int_{s^\prime \in \mathcal{S}} \Big(p(s^\prime | s_t, \pi_k(s_t), \theta^k) - p(s^\prime | s_t, \pi_k(s_t), \theta^*) \Big) h^{\pi_k}(s^\prime, \theta^k) d s^\prime \\
    &\leq \lrn{p(\cdot | s_t, \pi_k(s_t), \theta^k) - p(\cdot | s_t, \pi_k(s_t), \theta^*)} \lrn{h(\cdot, \theta^k)}_{\infty}  \\
    &= H L_p \lrn{\theta^* - \theta^k}.
\end{align*}
Plugging the result into the definition of $\Delta_{err}$ concludes the proof.
\end{proof}

The above regret decomposition in Lemma \ref{lem:mdp_regret_decomp} holds regardless of the approximate sampling methods being employed. To derive the final regret bounds, we discuss in the context of General Parmeteration and Simple parameterization respectively.

\subsubsection{General Parametrization}

The first term in Equation (\ref{eqn:mdp_regret_decom}) corresponds to the accumulating approximation error over the time horizon $T$ due to the use of approximate sampling method. Upper bounding this term relies on the statistical guarantee provided by the adopted approximate sampling method, which is the main novelty of LPSRL. In this section, we focus on the regret guarantee under the general parameterization. 

To maintain the sub-linear regret guarantee, the convergence guarantee provided by SGLD is required to effectively upper bound the approximation error in the first term of Lemma \ref{lem:mdp_regret_decomp}.

\begin{lemma}
\label{lem:mdp_W1_bound}
Suppose Assumptions \ref{ass:1}-\ref{ass:4} are satisfied. Under the general parameterization of MDP, by instantiating LPSRL with SGLD, it holds that for any $p \geq 2$,
\begin{equation}
    \sum_{k=1}^{K_T} T_k W_1 (\Tilde{\rho}_{t_k}, \rho_{t_k}) \leq \sqrt{\frac{24T(\log T + 1)}{m}} (d + \log Q + (32 + 8d \kappa^2)p)^{1/2}~.
\end{equation}
\end{lemma}

\begin{proof} [Proof of Lemma \ref{lem:mdp_W1_bound}]
By design of Algorithm~\ref{alg:TS_alg_MDP_double} and the convergence guarantee of SGLD in Theorem \ref{thm:sgld_convergence}, we have,
\begin{align*} 
 \sum_{k=1}^{K_T} T_k W_1 (\Tilde{\rho}_{t_k}, \rho_{t_k})
&\leq \sum_{k=1}^{\log T + 1} 2^{k-1} W_p (\Tilde{\rho}_{t_k}, \rho_{t_k})\\
&\leq \sum_{k=1}^{\log T  + 1} 2^{k-1} \sqrt{\frac{12}{2^{k-1} m}} (d + \log Q + (32 + 8d \kappa^2)p)^{1/2}\\
&= \sqrt{\frac{12}{ m}} (d + \log Q + (32 + 8d \kappa^2)p)^{1/2} \sum_{k=1}^{\log T  + 1} \sqrt{2^{k-1}}\\
&\leq \sqrt{\frac{24T(\log T + 1)}{m}} (d + \log Q + (32 + 8d \kappa^2)p)^{1/2}.
\end{align*}
Here, the first inequality follows from the fact that $W_p \geq W_q$ for any $p \geq q$. The second equality directly follows from Theorem \ref{thm:sgld_convergence}, and the last inequality follows from the Cauchy-Schwartz inequality.

\end{proof}


To further upper bound $\Delta_{err}$ in Lemma \ref{lem:mdp_regret_term3_bound} under the General Parameterization, we establish the following concentration guarantee provided by SGLD under the static doubling batching scheme adopted by LPSRL.

\begin{lemma} [Concentration of SGLD]
\label{lem:MDP_SGLD_concentration}
For any policy-switch $k \in [K_T]$, instantiating LPSRL with SGLD guarantees that
\[
    \mathbb{E} \Big[T_k\|\theta^* - \theta^k\|^2 \Big] \leq \frac{960d}{m} \log T.
\]
\end{lemma}

\begin{proof} [Proof of Lemma \ref{lem:MDP_SGLD_concentration}]

At time $t_k$, denote $\theta^{k,*}$ the parameter sampling from the true posterior $\rho_{t_k}$, and $n_{t_k}$ the total number of available observations. By the triangle inequality, 
\begin{equation*}
    \lrn{\theta^* - \theta^k}^2 \leq 3(\lrn{\theta^* - \theta^{k,*}}^2 + \lrn{\theta^{k,*} - \theta^k}^2).
\end{equation*} 
Taking expectation and multiplying both sides by $2^{k-1}$ yields,
\begin{equation}
\label{eqn:mdp_sgld_con_1}
    \expectation [T_k\|\theta^* - \theta^k\|^2] \leq 3 \expectation[T_k \|\theta^* - \theta^{k,*}\|^2]  + 3 \expectation[T_k \|\theta^{k,*} - \theta^k\|^2].
\end{equation}

Under the General parameterization, we follow Assumption A2 in \cite{theocharous2017posterior} to focus on the MDPs that have proper concentration, which suggests the true parameter $\theta^*$ and the mode of the posterior $\theta^{k, *}$ satisfies
\[
    \expectation \left[\|\theta^* - \theta^{k,*}\|^2 \right] \leq \frac{32d}{m n_{t_k}} \log T.
\]
We provide an example in Appendix \ref{app:mdp_example} to show this assumption can be easily satisfied in practice. Let $\tilde{D}^2 := \frac{32d}{m} \log T$, then by Theorem \ref{thm:sgld_convergence_app} adapted to the MDP setting (i.e. with a change in notation),
we have,
\[
  W_2^2(\widetilde{\rho}_{t_k}, \rho_{t_k}) \leq \frac{4\tilde{D}^2}{n_{t_k}}.
\]
Note that $n_{t_k} = \sum_{k^\prime = 1}^{k-1} T_{k^\prime}$ and by design of Algorithm~\ref{alg:TS_alg_MDP_double}, $n_{t_k} \leq T_k \leq 2 n_{t_k}$. Combining the above results and Equation (\ref{eqn:mdp_sgld_con_1}), we have 
\begin{align*}
    \expectation \left[T_k\|\theta^* - \theta^k\|^2 \right] \leq 30\tilde{D}^2 \leq \frac{960d}{m} \log T, ~~~~~\forall \tilde{D}^2 \geq \frac{32d}{m} \log T.
\end{align*}
\end{proof}



With all the above results, we are now ready to prove the main theorem for LPSRL with SGLD.

\MDPRegretSGLD*

\begin{proof}[Proof of Theorem \ref{thm:mdp_general}]

First we further upper bound $\Delta_{err}$ using the concentration guarantee provided by SGLD. 
We first note that by Cauchy–Schwarz inequality,
\begin{equation}
\label{eqn:mdp_delta_err_term_2}
    \sum_{k=1}^{K_T} \sum_{t = t_k}^ {t_{k+1}-1}  \lrn{\theta^* - \theta^k}
    = \sum_{t=1}^T \lrn{\theta^* - \theta^k} 
    \leq \sqrt{T \sum_{t=1}^T \lrn{\theta^* - \theta^k}^2}
    =\sqrt{T \sum_{k=1}^{K_T} T_k \lrn{\theta^* - \theta^k}^2}.
\end{equation}
Combining Equation (\ref{eqn:mdp_delta_err_term_1}) in Lemma \ref{lem:mdp_regret_term3_bound} and (\ref{eqn:mdp_delta_err_term_2}), by Theorem~\ref{thm:mdp_communication_cost},
Lemma \ref{lem:mdp_regret_term3_bound} and \ref{lem:MDP_SGLD_concentration}, we have, 
\begin{align*}
    \Delta_{err}
    &\leq H L_p \sqrt{T \mathbb{E} \Big[ \sum_{k=1}^{K_T} T_k \lrn{\theta^* - \theta^k}^2 \Big] } \\
    &\leq H L_p \sqrt{T K_T \max_k \mathbb{E}  \Big[ T_k  \lrn{\theta^* - \theta^k}^2 \Big] } \\
    &\leq H L_p \sqrt{\frac{960d}{m}T log T (\log T + 1)} \\
   &\leq H (\log T + 1) \sqrt{\frac{960d}{m}T} \numberthis \label{eqn:mdp_err_bound}.
\end{align*}

Then combining Lemma \ref{lem:mdp_regret_decomp}, \ref{lem:mdp_W1_bound} and Equation \ref{eqn:mdp_err_bound}, we have,
\begin{align*}
    R_B(T) 
    &\leq H(\log T + 1) + H (\log T + 1) \sqrt{\frac{960d}{m}T} + \sqrt{\frac{24T(\log T + 1)}{m}} (d + \log Q + (32 + 8d \kappa^2)p)^{1/2} \\
    &\leq (1 + \sqrt{960} +\sqrt{24})H (\log T + 1) \sqrt{\frac{T}{m}} (d + \log Q + (32 + 8d \kappa^2)p)^{1/2} \\
    & \leq 38 H (\log T + 1) \sqrt{\frac{T}{m}} (d + \log Q + (32 + 8d \kappa^2)p)^{1/2}.
\end{align*}
\end{proof}

\subsubsection{Simplex Parametrization}
We now discuss the performance of LPSRL under simplex parametrization. 
Similar to the General Parameterization, the regret guarantee of LPSRL relies on the convergence guarantee of MLD, which is presented in the following theorem.

\MLDConvergence*

\begin{proof}[Proof of Theorem \ref{thm:mdp_mld}]
Theorem \ref{thm:mdp_mld} follows from Theorems 2 and 3 from \citet{hsieh2018mirrored} with step sizes given as per Theorem 3 from \citet{cheng2018convergence}.
\end{proof}

Instantiating Algorithm~\ref{alg:TS_alg_MDP_double} with MLD provides the following statistical guarantee to control the approximation error in terms of the Wasserstein-$1$ distance.

\begin{lemma}
\label{lem:mdp_tabular_W1_bound}
Under the simplex parameterization of MDPs, we run MLD for each state-action pair $(s, a) \in \states \times \actions$ at the beginning of each policy-switch $k \in [K_T]$ for $\tilde{O}(|\states||\actions| n_{t_k})$ iterations. Suppose Assumption \ref{ass:mdp_lip_J} and \ref{ass:mdp_lip_transition} are satisfied, then by instantiating LPSRL (Algorithm \ref{alg:TS_alg_MDP_double}) with MLD (Algorithm \ref{alg:TS_MLD}) as \texttt{SamplingAlg}, we have,
\begin{equation}
   \sum_{k=1}^{K_T} T_k W_1(\tilde{\rho}_{t_k}, \rho_{t_k}) \leq |\states| \sqrt{8 |\actions| T\log T}.
\end{equation}
\end{lemma}

\begin{proof} [Proof of Lemma~\ref{lem:mdp_tabular_W1_bound}]
By Corollary~\ref{cor:MDP_tabular_PS_main_change}, in tabular settings, the error term in the Wasserstein-$1$ distance can be further decomposed in terms of state-action pairs, suggesting
\[
    W_1(\tilde{\rho}_{t_k}, \rho_{t_k}) = \sum_{(s,a) \in \mathcal{S} \times \mathcal{A}} W_1 (\Tilde{\rho}_{t_k}(s,a), \rho_{t_k}(s,a)).
\]
    
Then by design of Algorithm~\ref{alg:TS_alg_MDP_double}, we have,
\begin{align*}
    \sum_{k=1}^{K_T} T_k W_1(\tilde{\rho}_{t_k}, \rho_{t_k})
    &=\sum_{k=1}^{K_T} T_k \sum_{s,a} W_1(\tilde{\rho}_{t_k}(s,a), \rho_{t_k}(s,a)) \\
    &\leq \sum_{k=1}^{\log T + 1}  T_k \sum_{s,a} W_2(\tilde{\rho}_{t_k}(s,a), \rho_{t_k}(s,a)) \\
    &\leq \sum_{k=1}^{\log T + 1}  T_k |\states||\actions| \max_{s,a} W_2(\tilde{\rho}_{t_k}(s,a), \rho_{t_k}(s,a)) \numberthis \label{eqn:mdp_w1_eqn1},
\end{align*}
where the first inequality follows from the fact that $W_p \geq W_q$ for any $p \geq q$.

The convergence guarantee provided by Theorem \ref{thm:mdp_mld} for MLD suggests, for each state-action pair $(s, a) \in \states \times \actions$, upon running MLD for $\tilde{O}(|\states||\actions| n_{t_k})$ iterations, where $n_{t_k}$ is the number of data available for $(s, a)$ at time $t_k$, we have
\begin{equation}
\label{eqn:mdp_w1_eqn2}
    W_2 (\tilde{\rho}_{t_k}(s,a), \rho_{t_k}(s,a)) \leq \sqrt{\frac{1}{| \actions| n_{t_k}}}.
\end{equation}
At time $t_k$, let $n_{t_k}$ be the total number of available observations, which gives $n_{t_k} = \sum_{k^\prime = 1}^{k-1} T_{k^\prime}$ and $t_k = n_{t_k} + 1$. By design of Algorithm~\ref{alg:TS_alg_MDP_double}, $n_{t_k} \leq T_k \leq 2 n_{t_k}$. Then combining Equation (\ref{eqn:mdp_w1_eqn1}) and (\ref{eqn:mdp_w1_eqn2}) gives,
\begin{align*}
    \sum_{k=1}^{K_T} T_k \sum_{s,a} W_1(\tilde{\rho}_{t_k}(s,a), \rho_{t_k}(s,a))
    &\leq \sum_{k=1}^{\log T + 1} |\states|\sqrt{2 | \actions| T_k} \\
    &\leq \sum_{k=1}^{\log T + 1} |\states| \sqrt{ | \actions| 2^k} \\
    &\leq |\states| \sqrt{| \actions| (\log T + 1) \sum_{k=1}^{\log T + 1} 2^{k}} \\
    &\leq |\states| \sqrt{8 |\actions| T\log T},
\end{align*}
where the third inequality follows from the  Cauchy-Schwarz inequality.

\end{proof}

Lemma~\ref{lem:mdp_tabular_W1_bound} suggests the approximation error in the first term of Lemma~\ref{lem:mdp_regret_decomp} can be effectively bounded when instantiating \texttt{SamplingAlg} with MLD.

\begin{lemma} [Concentration of MLD] \label{lem:MDP_MLD_concentration}
For any policy-switch $k \in [K_T]$, we run MLD (Algorithm \ref{alg:TS_MLD}) for each state-action pair $(s, a) \in \states \times \actions$ at time $t_k$ for $\tilde{O}(|\states||\actions| n_{t_k})$ iterations. Then instantiating LPSRL with MLD guarantees that
\[
    \expectation [T_k \|\theta^{k, *} - \theta^k\|^2] \leq 2 |\states|^2 | \actions|.
\]
\end{lemma}

\begin{proof} [Proof of Lemma \ref{lem:MDP_MLD_concentration}]
By tower's rule and the triangle inequality, we have 
\begin{align*}
    \expectation \left[T_k \|\theta^{k, *} - \theta^k\|^2 \right] 
    & = \expectation \left[ \expectation \left[T_k \|\theta^{k, *} - \theta^k\|^2 \right] \Big | \mathcal{H}_{t_k} \right] \\
    &\leq \expectation \left[ T_k W_2^2(\tilde{\rho}_{t_k}, \rho_{t_k}) \Big | \mathcal{H}_{t_k} \right] \\
    &\leq \expectation \left[ T_k (|\states||\actions|)^2 \max_{s,a} W_2^2 (\tilde{\rho}_{t_k}(s,a), \rho_{t_k}(s,a)) \Big | \mathcal{H}_{t_k} \right] \numberthis \label{eqn:mdp_w1_eqn3}.
\end{align*}
where the last inequality follows from the fact that in tabular setting,  \resizebox{0.38\hsize}{!}{$ W_2^2(\tilde{\rho}_{t_k}, \rho_{t_k}) = \left (\sum_{s,a}W_2 (\tilde{\rho}_{t_k}(s,a), \rho_{t_k}(s,a)) \right)^2$}.

By the convergence guarantee of MLD in Theorem \ref{thm:mdp_mld}, for each state-action pair $(s, a) \in \states \times \actions$, upon running MLD for $\tilde{O}(|\states||\actions| n_{t_k})$ iterations, we have
\begin{equation}
\label{eqn:mdp_w1_eqn4}
    W_2 (\tilde{\rho}_{t_k}(s,a), \rho_{t_k}(s,a)) \leq \sqrt{\frac{1}{| \actions| n_{t_k}}}.
\end{equation}
Combining Equation (\ref{eqn:mdp_w1_eqn3}) and (\ref{eqn:mdp_w1_eqn4}) and the fact that $n_{t_k} \leq T_k \leq 2 n_{t_k}$ concludes the proof.


\end{proof}

With the concentration guarantee between sample $\theta^k$ and $\theta^{k, *}$, as well as the concentration guarantee between $\theta^{k, *}$ and $\theta^{*}$ in exact PSRL, we are able to effectively upper bound the model estimation error $\Delta_{err}$ in tabular settings.

\begin{lemma} [Bound $\Delta_{err}$ in tabular settings]
\label{lem:mdp_mld_delta}
With the definition of model estimation error  $\Delta_{err}$ in Lemma \ref{lem:mdp_regret_term3_bound}, in tabular setting, the following upper bound holds for $\Delta_{err}$,
\begin{equation}
    \Delta_{err} \leq 66 H |\states| \sqrt{|\actions| T \log(2|\states||\actions|T) },
\end{equation}
where $L_p$ is the Lipschitz constant for transition dynamics.
\end{lemma}
\begin{proof} [Proof of Lemma \ref{lem:mdp_mld_delta}]
By the triangle inequality and Lemma \ref{lem:mdp_regret_term3_bound},
\begin{equation}
\label{eqn:mdp_mld_delta_1}
    \Delta_{err} \leq H L_p \mathbb{E} \Big[ \sum_{k=1}^{K_T} \sum_{t = t_k}^ {t_{k+1}-1}  \lrn{\theta^* - \theta^k} \Big] \leq H L_p \left( \mathbb{E} \Big[ \sum_{k=1}^{K_T} \sum_{t = t_k}^ {t_{k+1}-1}  \lrn{\theta^* - \theta^{k, *} } \Big] + \mathbb{E} \Big[ \sum_{k=1}^{K_T} \sum_{t = t_k}^ {t_{k+1}-1}  \lrn{\theta^{k, *} - \theta^k}\Big] \right).
\end{equation}

\textbf{Bound The first term.} The first term can be upper bounded using standard concentration results of exact PSRL algorithms in Bayesian settings. Define the event 
\begin{equation} 
    E_\theta = \left\{\theta: \forall (s, a) \in \states \times \actions, ~~~~ \lrn{\theta(\cdot | s,a) - \widehat{\theta}^{k}(\cdot | s,a)}_1 \leq \beta_k(s, a) \right\},
\end{equation}
where $\widehat{\theta}^{k}$ is the empirical distribution at the beginning of policy switch  $k$, $\beta_k(s, a) := \sqrt{\frac{14|\states|\log(2|\states||\actions|t_k T)}{max(1, n_{t_k}(s, a))}}$ following \cite{jaksch2010near, ouyang2017learning, osband2013more} by setting $\delta = 1 / T$. Then event $E_\theta$ happens with probability at least $1-\delta$. Note that for any vector x, $\lrn{x}_2 \leq \lrn{x}_1$, and by the triangle inequality, we have
\[
    \lrn{\theta^* - \theta^{k, *} } \leq \sum_{s^\prime \in \states} \Big|\theta^*(\cdot | s,a) - \theta^{k, *}(\cdot | s,a) \Big| \leq 2(\beta_k(s_t, a_t) + \mathbf{1}_{\{\theta^{*} \notin E_\theta\}} ).
\]
At any time $t \in [t_k, t_k + T_k -1]$, $n_t \leq 2 n_{t_k}$ for any state-action pair $(s_t, a_t)$, and by the fact that $t_k \leq T$, we have
\begin{align}
\label{eqn:mdp_mld_delta_3}
    \mathbb{E} \Big[ \sum_{k=1}^{K_T} \sum_{t = t_k}^ {t_{k+1}-1} \beta_k(s_t, a_t) \Big] \leq \sum_{k=1}^{K_T} \sum_{t = t_k}^ {t_{k+1}-1} \sqrt{\frac{28 |\states|\log(2|\states||\actions|t_k T)}{max(1, n_{t}(s_t, a_t))}} 
    \leq \sum_{t=1}^{T} \sqrt{\frac{56 |\states|\log(2|\states||\actions|T)}{max(1, n_{t}(s_t, a_t))}}.
\end{align}
It then suffices to bound $\sum_{t=1}^T 1 / \sqrt{max(1, n_{t}s_t, a_t)}$. Note that
\begin{align*}
    \sum_{t=1}^T \frac{1}{ \sqrt{max(1, n_{t}(s_t, a_t))}}
    &= \sum_{(s, a)} \sum_{t=1}^T \frac{\mathbf{1}_{(s_t, a_t)=(s, a)}}{ \sqrt{max(1, n_{t}(s, a))}}\\
    &\leq 4 \sum_{(s, a)} \int_{z = 0}^{n_{T+1}(s, a)} z^{-1/2} dz \\
    &\leq 4 \sqrt{|\states||\actions| \sum_{(s, a)}n_{T+1}(s,a)} \\
    &\leq 4 \sqrt{|\states||\actions|T}\numberthis \label{eqn:mdp_mld_delta_term_1}.
\end{align*}
On the other hand, by definition of $\beta_k(s,a)$, $\mathbb{P}(\theta^{*} \notin E_\theta\}) \leq 1 / (T t_k^6)$, which yields
\begin{align}
\label{eqn:mdp_mld_delta_term_3}
    \mathbb{E} \left[ \sum_{k=1}^{K_T} \sum_{t = t_k}^ {t_{k+1}-1} \mathbf{1}_{\{\theta^{*} \notin E_\theta\}} \right] \leq \mathbb{E} \left[ \sum_{k=1}^{K_T} T_k \mathbb{P}(\theta^{*} \notin E_\theta\}) \right] 
    \leq \sum_{k=1}^{\infty} k^{-6} \leq \sum_{k=1}^{\infty} k^{-2} \leq 2.
\end{align}
Combining Equation (\ref{eqn:mdp_mld_delta_3}), 
(\ref{eqn:mdp_mld_delta_term_1}) and (\ref{eqn:mdp_mld_delta_term_3}), we have,
\begin{equation}
\label{eqn:mdp_mld_delta_term_4}
     H L_p \mathbb{E} \Big[ \sum_{k=1}^{K_T} \sum_{t = t_k}^ {t_{k+1}-1}  \lrn{\theta^* - \theta^{k, *} } \Big] \leq 64 H L_p |\states| \sqrt{|\actions| T \log(2|\states||\actions|T) }
\end{equation}

\textbf{Bound the second term.} The second term arises from the use of approximate sampling. Note that by Cauchy–Schwarz inequality, this term in Equation~(\ref{eqn:mdp_mld_delta_1}) satisfies,
\begin{equation}
\label{eqn:mdp_mld_delta_term_2}
    \sum_{k=1}^{K_T} \sum_{t = t_k}^ {t_{k+1}-1}  \lrn{\theta^{k, *} - \theta^k}
    = \sum_{t=1}^T \lrn{\theta^{k, *} - \theta^k} 
    \leq \sqrt{T \sum_{t=1}^T \lrn{\theta^{k, *} - \theta^k}^2}
    =\sqrt{T \sum_{k=1}^{K_T} T_k \lrn{\theta^{k, *} - \theta^k}^2}.
\end{equation}
It then relies on the concentration guarantee provided by MLD for LPSRL under the static policy switch scheme. By Lemma \ref{lem:MDP_MLD_concentration}, we have, 
\begin{align*}
    &H L_p \sqrt{T \mathbb{E} \Big[ \sum_{k=1}^{K_T} T_k \lrn{\theta^* - \theta^k}^2 \Big] } \leq H L_p \sqrt{T K_T \max_k \mathbb{E}  \Big[ T_k  \lrn{\theta^* - \theta^k}^2 \Big] } \leq H L_p |\states| \sqrt{4 | \actions| T \log T} \numberthis \label{eqn:mdp_mld_delta_term_5}.
\end{align*}

Combining Equation (\ref{eqn:mdp_mld_delta_term_4}) and (\ref{eqn:mdp_mld_delta_term_5}) concludes the proof.

\end{proof}
 
With all the above results, we now proceed to prove the regret bound for LPSRL with MLD.

\MDPRegretMLD*

\begin{proof}[Proof of Theorem \ref{thm:mdp_tabular_main}]

By Lemma~\ref{lem:mdp_regret_decomp},  \ref{lem:mdp_tabular_W1_bound} and \ref{lem:mdp_mld_delta}, we have
\begin{align*}
    R_B(T) 
    &\leq H(\log T + 1) + 66 H |\states| \sqrt{|\actions| T \log(2|\states||\actions|T) } + |\states| \sqrt{8 |\actions| T\log T} \\
    &\leq 2H \log T + 66 H |\states| \sqrt{|\actions| T \log(2|\states||\actions|T) } + 4 |\states| \sqrt{ | \actions| T\log T} \\
    & \leq 72 H |\states| \sqrt{|\actions| T \log(2|\states||\actions|T) }.
\end{align*}
By Lemma \ref{lem:mdp_tabular_W1_bound}, for each state-action pair $(s, a) \in \states \times \actions$ and policy-switch $k \in [K_T]$, the number of iterations required for MLD is $O(|\states||\actions| 2^{k-1})$. This suggests  that for each state-action pair, the total number of iterations required for MLD is $O(|\states||\actions| T)$ along the time horizon $T$. Summing over all possible state-action pairs, the computational cost of running MLD in terms of the total number of iterations is $O(|\states|^2|\actions|^2 T)$.
\end{proof}

\subsection{General Parameterization Example}
\label{app:mdp_example}

Following \cite{theocharous17poi, theocharous2017posterior} we consider a points of interest (POI) recommender system where the system recommends a sequence of points that could be of interest to a particular tourist or individual. We will let the points of interest be denoted by points on $\reals$. Following the perturbation model in \cite{theocharous17poi, theocharous2017posterior}, the transition probabilities are $p(s|\theta) = p(s)^{1/\theta}$ if the chosen action is $s$ and it is $p(s)/z(\theta)$ otherwise. Here $s$ is a state or a POI and $z(\theta) = \frac{\sum_{x \neq s} p(x)}{1-p(s)^{1/\theta}}$. Furthermore, to fully specify $p$ we consider $p(s|\theta) = \frac{1}{\sqrt{2\pi}}e^{-s^2/2\theta} $. One can see that Assumptions $1$-$4$ are satisfied due to the Gaussian-like nature of the transition dynamics and the satisfiability of Assumption $5$ follows from Lemma 5 in \cite{theocharous2017posterior}.

\section{Experimental Details}
\label{sec:appendix_empirical}

\subsection{Additional Discussions of Langevin TS in Gaussian Bandits}
\begin{figure}[!ht]
    \centering
    \begin{subfigure}{0.49\linewidth}
        \centering
        \includegraphics[width=\linewidth]{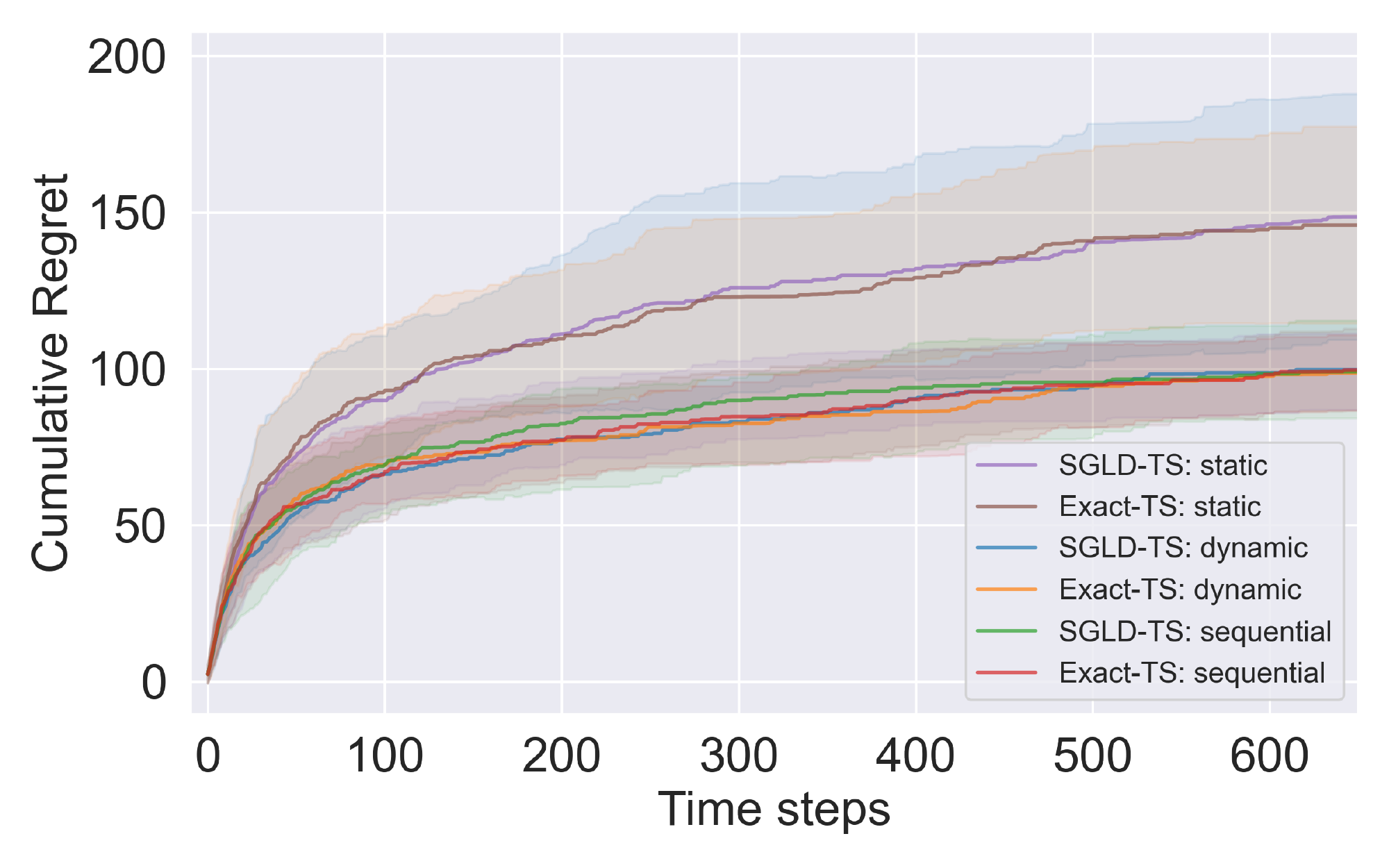}
    \end{subfigure}
    \begin{subfigure}{0.49\linewidth}
        \centering
        \includegraphics[width=\linewidth]{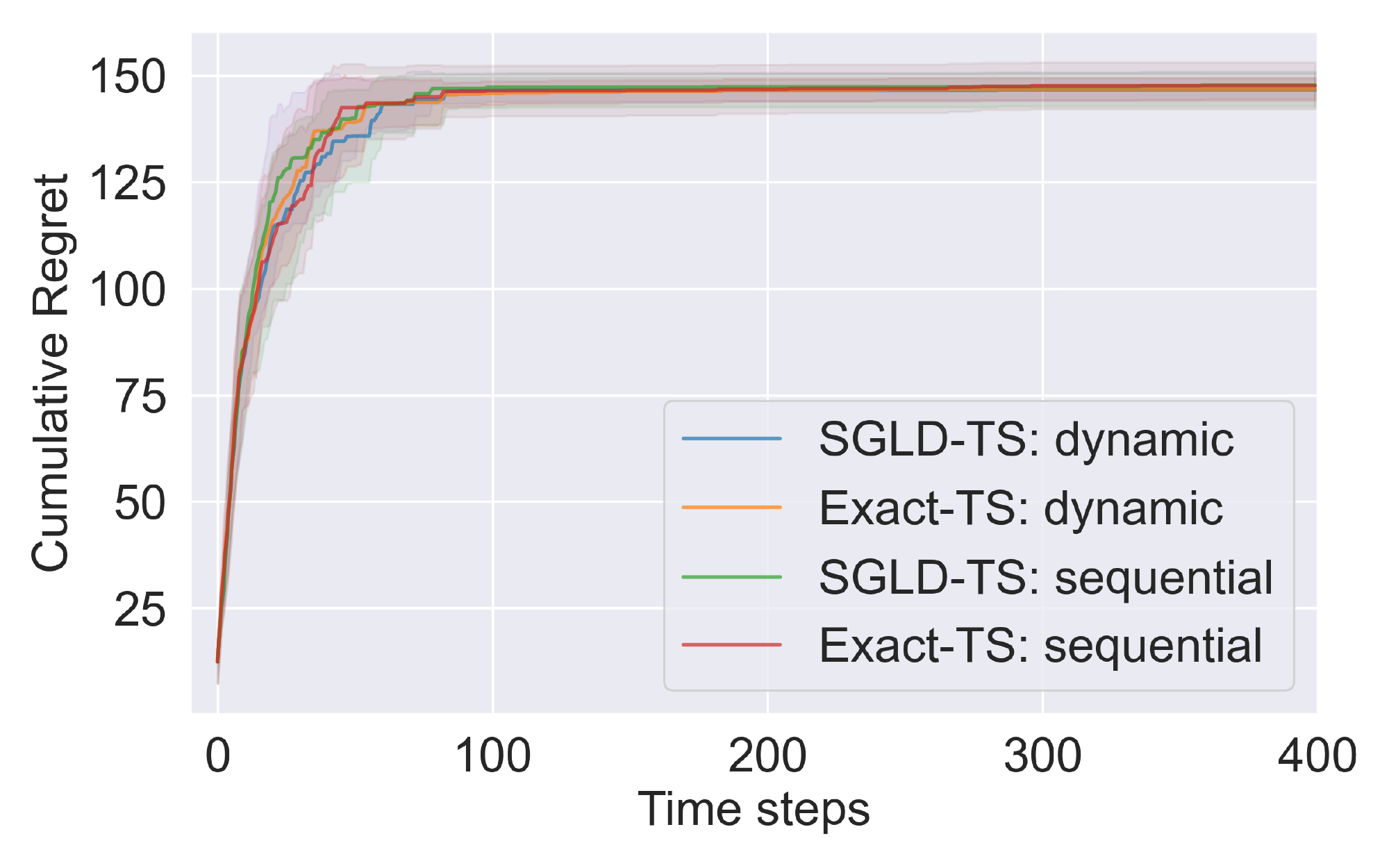}
    \end{subfigure}
    \caption{Left:(a) Expected regret for informative priors. Right:(b) Expected regret for uninformative priors. Results are reported over 10 experiments. In both scenarios, SGLD-TS under dynamic scheme achieves optimal performance as in sequential case without using approximate sampling.}
    \label{fig:bandit_regret_2}
\end{figure} 

\begin{figure}[!ht]
    \centering
    \begin{subfigure}{0.49\linewidth}
        \centering
        \includegraphics[width=\linewidth]{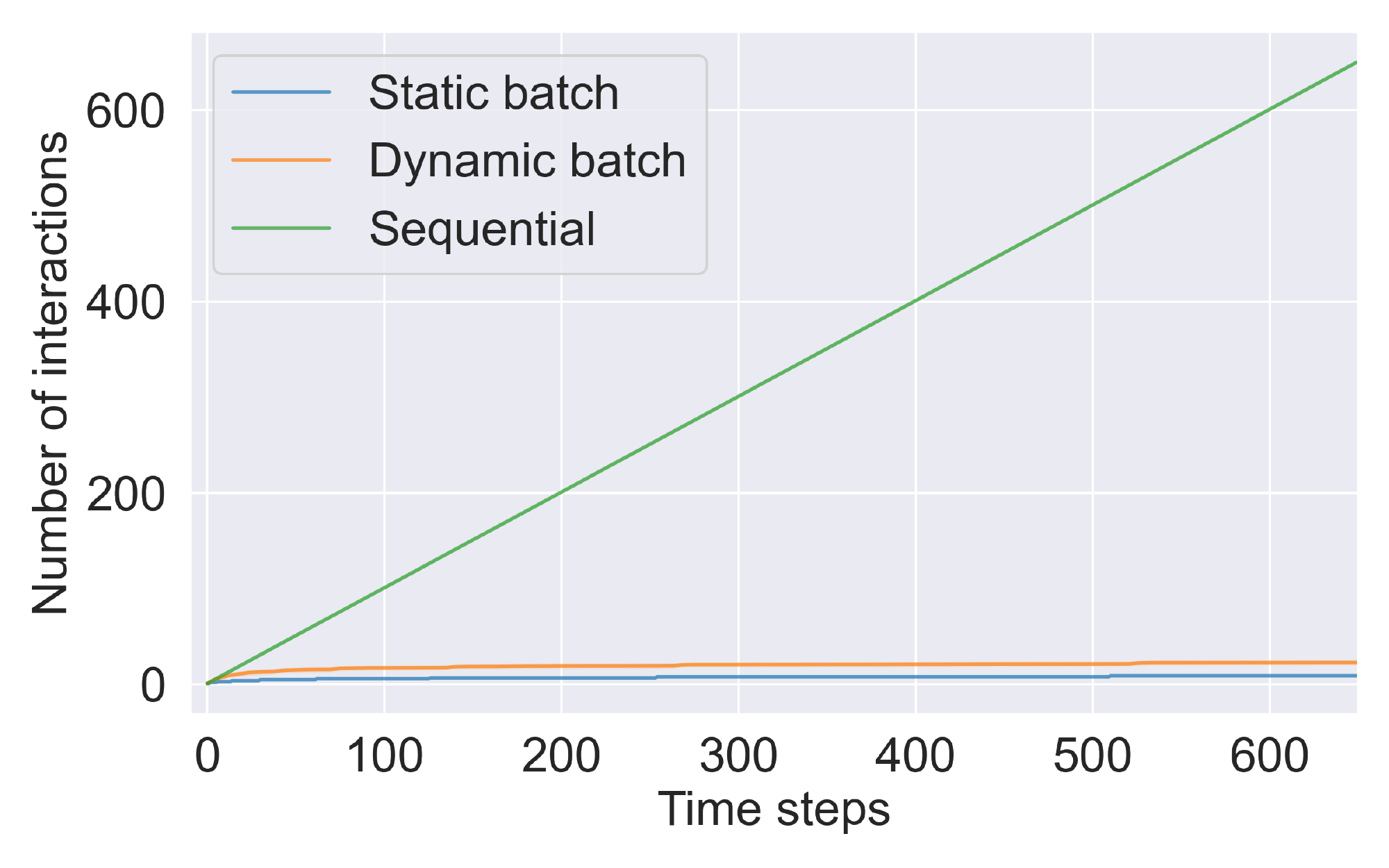}
    \end{subfigure}
    \begin{subfigure}{0.49\linewidth}
        \centering
        \includegraphics[width=\linewidth]{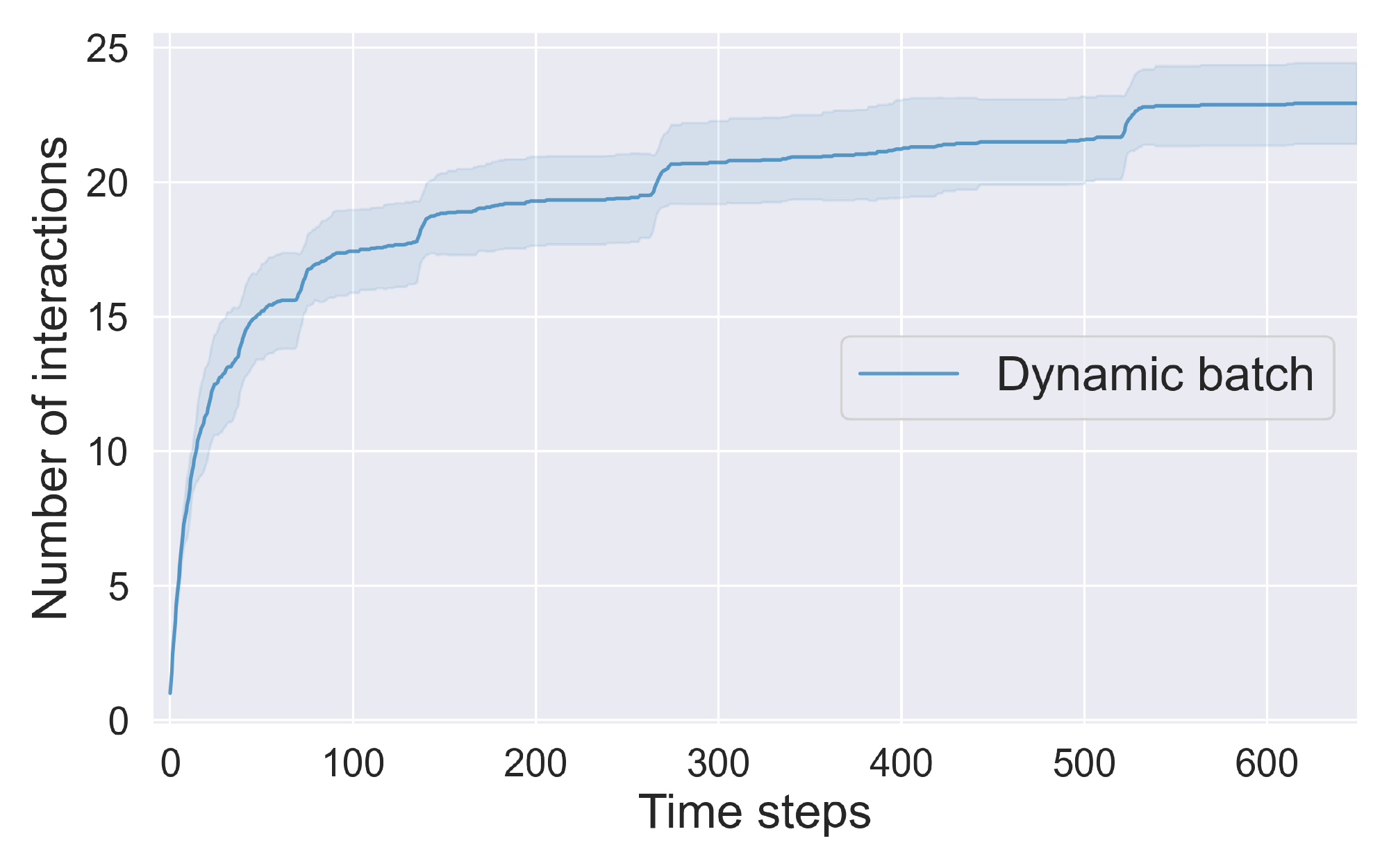}
    \end{subfigure}
    \caption{Left:(a)~expected communication cost of three batching schemes: fully-sequential mode, dynamic batch, and static batch. Right:(b)~expected communication cost under dynamic batching scheme for Gaussian bandits.}
    \label{fig:bandit_communication}
\end{figure} 

In this section, we present additional empirical results for the Gaussian bandit experiments. In particular, we examine both informative priors and uninformative priors for Gaussian bandits with $N=15$ arms, where each arm is associated with distinct expected rewards. We set the true expected rewards of all arms to be evenly spaced in the interval $[1, 20]$, and the ordering of values is shuffled before assigning to arms. All arms share the same standard deviation of $0.5$. We investigate the performance of SGLD-TS against UCB1, Bayes-UCB, and exact-TS under different interaction schemes: fully-sequential mode, dynamic batch scheme, and static batch scheme.

In the first setting, we assume prior knowledge of the ordering of expected rewards and apply informative priors to facilitate the learning process. Gaussian priors are adopted with means evenly spaced in $[14, 20]$, and inverted variance (i.e., precision) set to $0.375$. The priors are assigned according to the ordering of the true reward distributions. Note that the exact knowledge of the true expected values is not required. In TS algorithms, the selection of arms at each time step is based on sampled values, therefore efficient learning is essential even with the knowledge of the correct ordering. The expected regret of all methods is reported over 10 experiments and results are illustrated in Figure~\ref{fig:bandit_regret_2}(a). Results of both Figure~\ref{fig:regret}(a) and Figure~\ref{fig:bandit_regret_2}(a) demonstrate that SGLD-TS achieves optimal performance similar to exact-TS with conjugate families. Its appealing empirical performance in comparison to other popular methods (e.g., UCB1 and Bayes-UCB), along with its ability to handle complex posteriors using MCMC algorithms, make it a promising solution for challenging problem domains. Additionally, the introduction of the dynamic batch scheme ensures the computational efficiency of SGLD-TS. As depicted in Figure~\ref{fig:bandit_communication}(a)(b) and Table \ref{tab:exp_bandit_reg} (column labeled "batches"), communication cost is significantly reduced from linear to logarithmic dependence on the time horizon, as suggested by Theorem \ref{thm:bandit_log_batch}. Furthermore, in bandit environments, our dynamic batch scheme exhibits greater robustness compared to the static batch scheme for both frequentist and Bayesian methods.

Furthermore, we explore the setting where prior information is absent, and uninformative priors are employed. In this case, we adopt the same Gaussian priors as $\mathcal{N}(14.0, 8.0)$ for all arms. Similar to the first setting, the same conclusion can be drawn for SGLD-TS from Figure~\ref{fig:bandit_regret_2}(b).

\subsection{Experimental Setup for Langevin TS in Laplace Bandits}

In order to demonstrate the performance of Langevin TS in a broader class of general bandit environments where closed-form posteriors are not available and exact TS is not applicable, we construct a Laplace bandit environment consisting of $N = 10$ arms. Specifically, we set the expected rewards to be evenly spaced in the interval $[1, 10]$, and shuffle the ordering before assigning each arm a value. The reward distribution of each arm shares the same standard deviation of $0.8$. We adopt favorable priors to incorporate the knowledge of the true ordering in Laplace bandits. It is important to note that our objective is to learn the expected rewards, and arm selection at each time step is based on the sampled values rather than the ordering. In particular, we adopt Gaussian priors with means evenly spaced in $[4, 10]$ (ordered according to prior knowledge). The inverted variance (i.e., precision) for all Gaussian priors is set to $0.875$. We conduct the experiments 10 times and report the cumulative regrets in Figure \ref{fig:regret}(b).

By employing Langevin TS in the Laplace bandit environment, we aim to showcase the algorithm's effectiveness and versatility in scenarios where posteriors are intractable and exact TS cannot be directly applied. 

\begin{figure}[!ht]
    \centering
    \begin{subfigure}{0.49\linewidth}
        \centering
        \includegraphics[width=\linewidth]{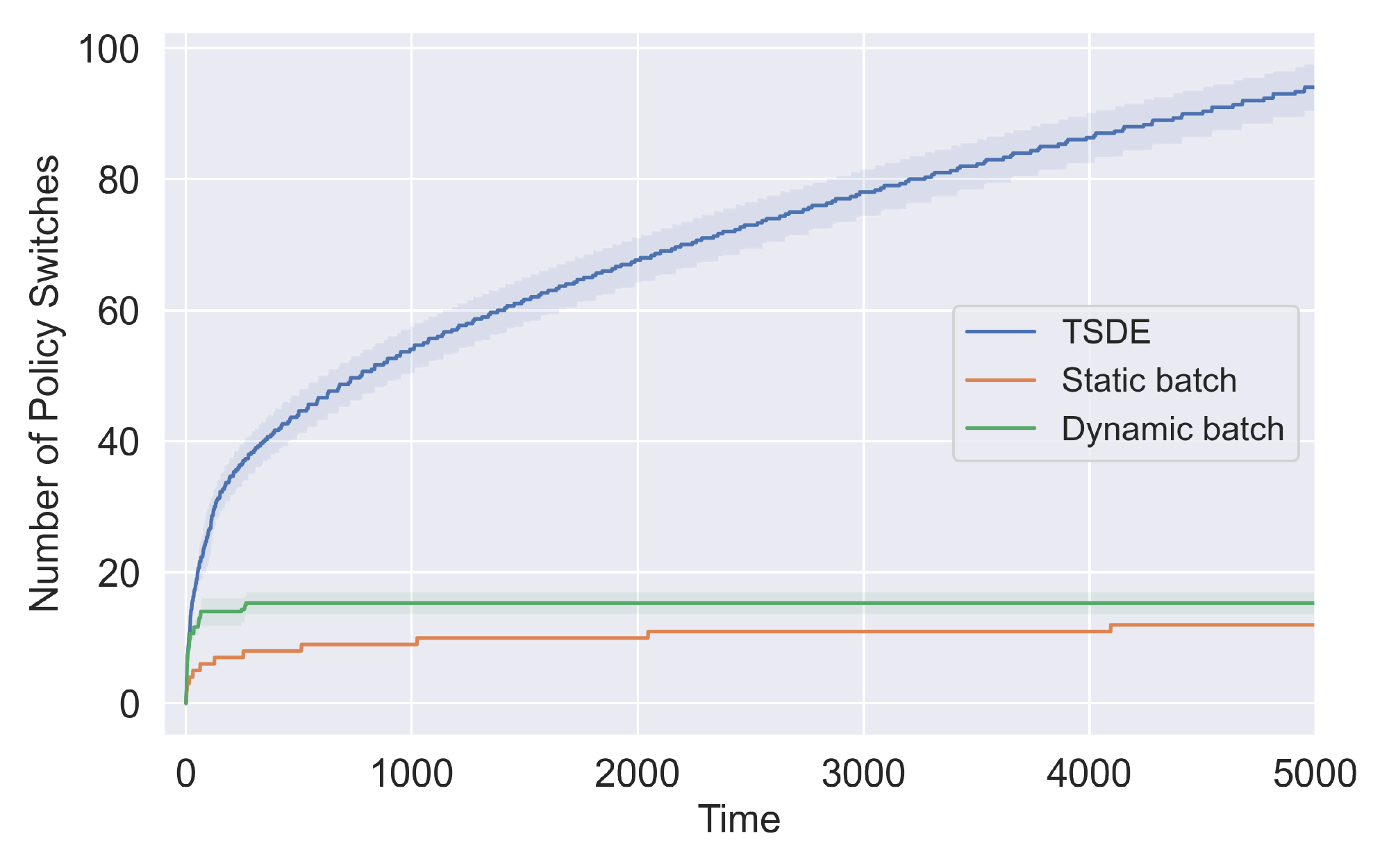}
    \end{subfigure}
    \caption{Number of policy switches in RiverSwim over 10 experiments. Static policy switch scheme requires the least number of communication.}
    \label{fig:MDP_policy_switches}
\end{figure}

\subsection{Experimental Setup for Langevin PSRL}

In MDP setting, we consider a variant of RiverSwim environment being frequently used empirically \citep{strehl2008analysis}, in which the agent swimming in the river is modeled with five states, and two available actions: left and right. If the agent swims rightwards along the river current, the attempt to transit to the right is going to succeed with a large probability of $p=0.8$. If the agent swims leftwards against the current, the transition probability to the left is small with $p=0.2$. Rewards are zero unless the agent is in the leftmost state ($r = 2.0$) or the rightmost state ($r = 10.0$). The agent is assumed to start from the leftmost state. We implement MLD-PSRL and exact-PSRL under two policy switch schemes, one is the static doubling scheme discussed in section \ref{sec:mdp}, and the other is the dynamic doubling scheme based on the visiting counts of state-action pairs. To ensure the performance of TSDE, we adopt its original policy switch criteria based on the linear growth restriction on episode length and dynamic doubling scheme. We run experiments $10$ times, and report the average rewards of each method in Figure \ref{fig:regret}(c). The number of policy switches under different schemes is depicted in Figure \ref{fig:MDP_policy_switches}.

\end{document}